\documentclass[twoside,11pt]{article}

\usepackage{jmlr2e}
\usepackage{microtype}
\usepackage{graphicx}
\usepackage{subfigure}
\usepackage{booktabs} 
\usepackage{bm}
\usepackage{amsmath}
\usepackage{amsfonts}
\usepackage{amssymb}
\usepackage{color}
\usepackage{pdfpages}
\usepackage{float}
\usepackage{natbib}
\usepackage{algorithm}
\usepackage{algorithmic}
\usepackage{enumitem}
\usepackage{cancel}
\usepackage{changes}
\usepackage[export]{adjustbox}
\newcommand{\stkout}[1]{\ifmmode\text{\sout{\ensuremath{#1}}}\else\sout{#1}\fi}
\setdeletedmarkup{\stkout{#1}}
\setcitestyle{authoryear, open={(},close={)}}

\def\D{{\mathbf D}}

\def\X{{\mathbf X}}

\def\K{{\mathbf K}}
\def\Z{{\mathbf Z}}

\def\x{{\mathbf x}}

\def\z{{\mathbf z}}

\def\0{{\mathbf 0}}

\jmlrheading{ }{2019}{XX-XX}{XX/XX}{XX/XX}{xxxxxx}{XXXXX}

\ShortHeadings{Hierarchical Change-Point Detection}{Moreno-Mu\~noz, Ram\'irez and Art\'es-Rodr\'iguez}
\firstpageno{1}

\begin{document}

\title{Change-Point Detection on Hierarchical Circadian Models}

\author{\name Pablo Moreno-Mu\~noz \email pmoreno@tsc.uc3m.es \\
		\name David Ram\'irez \email david.ramirez@uc3m.es\\ 
		\name Antonio Art\'es-Rodr\'iguez \email antonio@tsc.uc3m.es\\ 
       \addr Department of Signal Theory and Communications\\
       Universidad Carlos III de Madrid\\
       Legan\'es, Madrid, Spain}


\maketitle   

\begin{abstract}
This paper addresses the problem of change-point detection on sequences of high-dimensional and heterogeneous observations, which also possess a periodic temporal structure. Due to the dimensionality problem, when the time between change-points is on the order of the dimension of the model parameters, drifts in the underlying distribution can be misidentified as changes. To overcome this limitation, we assume that the observations lie in a lower-dimensional manifold that admits a latent variable representation. In particular, we propose a hierarchical model that is computationally feasible, widely applicable to heterogeneous data and robust to missing instances. Additionally, the observations' periodic dependencies are captured by non-stationary periodic covariance functions. The proposed technique is particularly fitted to (and motivated by) the problem of detecting changes in human behavior using smartphones and its application to relapse detection in psychiatric patients. Finally, we validate the technique on synthetic examples and we demonstrate its utility in the detection of behavioral changes using real data acquired by smartphones.
\end{abstract}

\begin{keywords}
  Change-point detection, circadian models, heterogeneous data, latent variable models, non-stationary periodic covariance functions.
\end{keywords}

\section{Introduction}
\label{sec:intro}

Change-point detection (CPD) consists of locating abrupt transitions in the generative model of a sequence of observations. 
Detecting these abrupt transitions or change points (CPs) has been long studied and appears in a vast amount of real-world scenarios. For instance, the detection of change points in finance, where they are known as structural breaks, is of utmost importance since these changes could render the investment strategies useless \citep{andersson2006some,berkes2004sequential}. CPD is also a crucial for the analysis of social networks \citep{raginsky2012sequential,krishnamurthy2012quickest}, where it allows for the detection of appearing/disappearing relationships among the agents of the network. In cognitive radio \citep{mitola},  the detection of change points in the received signals corresponds to the start or stop of a transmission, which is required to decide whether the corresponding frequency band is available or not for its use by secondary users \citep{arts15,du15}. 

Due to its importance, CPD has been studied extensively since the seminal works by \citet{shewhart1931} and \citet{wald1945}. However, a theoretical analysis was not done until a few decades later with the works by \citet{page1955,shiryaev1963,lorden1971}. The problem of CPD can be faced from a statistical point of view, which typically requires a model defined a priori, and alternatively, model-free methods. Regarding the statistical approaches, they can be further divided between frequentist and Bayesian techniques. 

The main idea behind frequentist methods is to derive a measure, usually based on likelihood ratio tests \citep{page1955,lorden1971} or, more recently, on regularized likelihood ratios \citep{haynes2014,killick2012}, between the pre-change and post-change distributions and compare it with a threshold. On the other hand, Bayesian approaches are based on assigning a prior distribution over the change points, or a proper surrogate, and derive its posterior distribution using a probabilistic framework \citep{Basseville1993}. Typically, these last models focus on batch settings \citep{fearnhead2006exact,harchaoui2009kernel}, where the number of changes is often unknown. In contrast, online methods have been proposed based on particle filters \citep{fearnhead2007line} or the Bayesian online change-point detection (BOCPD) model of \citet{adams2007bayesian}. Online methods are particularly useful whenever a retrospective approach is not feasible due to the sequential availability of data samples. These works focus on inferring the posterior distribution of change points for univariate time series, which also allows to obtain maximum-a-posteriori (MAP) estimates thereof. An alternative approach within the Bayesian framework is to formulate the problem as a Markov optimal stopping problem \citep{poor2009}, where dynamic programming algorithms \citep{bertsekas2005,puterman2014} can be applied to compute optimal partitions.

In this work we focus on an application of CPD different from the aforementioned ones. We consider the problem of modeling and detecting changes in human behavior, where we believe this sort of methods is potentially sound but have not been explored yet. The detection of changes in human habits have many potential applications. To name a few, in customer modeling as a target of commercial marketing \citep{newman1957}, information security \citep{vroom2004} and surveillance \citep{cristani2013,gowsikhaa2014}, voting prediction \citep{braha2017}  or just to discover unfamiliar patterns in some subset of population. An important application appears when we focus on medical scenarios, particularly in the field of psychiatry. CPD can be used to detect behavioral changes, which in patients with prevalent chronic disorders (e.\,g., schizophrenia or chronic depression) may be signals of future relapses \citep{barnett2018,berrouiguet2018}. To monitor psychiatric patients' symptoms there exist several alternatives, where the smartphone-based monitoring stands out \citep{Strickland14}. Concretely, the ubiquitous presence of smartphones and the large amount of data gathered by these devices have opened new opportunities in this area. Among others, smartphones observations range from inertial sensors measurements or GPS information to timestamps and duration of calls or app usage log. However, all these monitoring traces pose some challenges that have not been considered in detail within the framework of change-point detection. First, the observations provided by smartphones are high-dimensional and heterogeneous. Moreover, since humans are strictly conditioned by $24$-hours periods, also known as circadian rhythms, samples usually present some underlying periodicities related to the human behavior \citep{pentland1999modeling,eagle2009eigenbehaviors}. Finally, since some of these sensors may fail or be turned off, it is common to have missing values in the data. 



A large portion of CPD algorithms proposed in the literature do not consider the aforementioned features of data captured by smartphones. Concretely, the general assumption of CPD models is to consider low-dimensional and homogenous datasets \citep{Basseville1993,Tartakovsky2015,adams2007bayesian,fearnhead2007line}, with a few exceptions that study multivariate data, such as \citet{xie2013change} and \citet{xuan2007modeling}. However, as mentioned above, in the detection of behavioral changes using smartphones, observations are usually high-dimensional and heterogeneous, i.e., they are composed by a mixture of continuous, categorical, binary or discrete variables. Heterogenous observations are difficult to deal with \citep{valera2017,nazabal2018,moreno2018}, since a not careful learning process would result in a system that mainly uses only one of the different data types.  For instance, one kind of observation which dominates an unbalanced likelihood. Moreover, the CPD problem becomes even more difficult when observations possess temporal structure, that is, they deviate from the independent and identically distributed (i.\,i.\,d.) assumption \citep{Basseville1993,Tartakovsky2015}. This is the novel case studied in this work, where there exists temporal structure induced by the circadian rhythm. The last of the aforementioned challenges, missing data, has also not been considered much in the past, with a few exceptions, such as \citet{xie2013change}.

In this paper, we address the problem using a Bayesian CPD approach. This sort of methods provide a general measure of uncertainty, which is critical for us, due to the need of reliable decision-making algorithms in the mental-health context. Particularly, there has been a considerable effort in Bayesian estimation of change points on different domains. For example, \citet{hohle2010online} introduced an online cumulative sum detector that finds changes on categorical time series. Other approaches are able to model multivariate real data using undirected Gaussian graphical models (GGMs) \citep{xuan2007modeling} or even mixture models \citep{alippi2015change,kuncheva2013change}, together with log-likelihood ratios or divergences.  

Here, we generalize the BOCPD algorithm \citep{adams2007bayesian} to address the challenges raised by the application of CPD in psychiatry. Several extensions of the BOCPD algorithm exist as well, including adaptive sequential methods \citep{turner2009adaptive} or the Gaussian Process BOCPD model \citep{saatcci2010gaussian}. This last work extends the BOCPD algorithm to locate change points from observations with arbitrary temporal structure. Additionally, unlike previous approaches, the non-exponential BOCPD model \citep{turner2013online} explored applications to new families of distributions, where computing posterior probabilities is intractable and variational inference is therefore required. Another  recent alternative  is \citet{agudelo2019}, which extends the BOCPD method to also predict the number of time steps until the next change point. 

The proposed generalization of the BOCPD algorithm considers the CPD problem on heterogeneous and high-dimensional sequences through hierarchical models within latent variables. In Section \ref{sec:hierarchical}, the hierarchical extension of the BOCPD method is described. First, we propose a general latent variable model and, later on, due to the complexity induced by arbitrary latent variables, we particularize it to categorical ones. This is also well justified by the proposed application since these categorical latent variables represent what we could call \emph{type of day}, i.e., one day could correspond to a typical workday, another one to a weekend, etc. Thus, what the proposed method does is detect changes in the distributions of such types of days. For instance, a change point should be detected if the category that represents typical workdays disappears. Moreover, for the hierarchical model we also present the inference procedure and two simplified models, as well as the approach to consider missing data. The association between observations and types of days is presented in Section \ref{sec:circadian}, where a clustering technique based on mixtures of Gaussians is described. Interestingly, the covariance matrix of the Gaussian random vectors will be designed to capture the circadian features induced by human behavior. We develop the inference based on the expectation-maximization (EM) algorithm \citep{dempster1977maximum}, which also allows us to handle missing data. Finally, the performance of the proposed method is evaluated in Section \ref{sec:experiments} by means of synthetic experiments and real data acquired by smartphone.

\section{Change-Point Detection on Hierarchical Models}
\label{sec:hierarchical}

This section generalizes the method proposed by \citet{adams2007bayesian} to consider hierarchical models, which allow us to deal with high-dimensional observations. \citet{adams2007bayesian} consider that the sequence of observations, $\mathbf{X}_{1:t}$, may be divided into non-overlapping partitions separated by change points. With a few exceptions \citep{saatcci2010gaussian}, the work in \citet{adams2007bayesian} and its generalizations assume that the data within each partition $\rho$ is i.\,i.\,d. according to some generative distribution $p(\mathbf{x}_{t}|\bm{\theta}_t)$ with unknown parameters $\bm{\theta}_t$. Hence, these parameters are constant within each partition but they change between change points,\footnote{Note that not every component of $\bm{\theta}_t$ needs to change between CPs.} i.e.,
\begin{equation}
  \bm{\theta}_t = \begin{cases}
  \bm{\theta}_a, & t < \text{CP}_1, \\
  \bm{\theta}_b, & \text{CP}_1 \leq t \leq \text{CP}_2, \\
  \bm{\theta}_c, & \text{CP}_2 \leq t \leq \text{CP}_3, \\
  & \vdots
  \end{cases}
\end{equation}
where $\text{CP}_i$ is the time index of $i$th change point and $\bm{\theta}_a \neq \bm{\theta}_b \neq \bm{\theta}_c = \cdots$. Moreover, the run length was defined in \citet{adams2007bayesian} as an auxiliary variable to denote when change points appear:
\begin{equation}
  r_t = \begin{cases} 0, & \text{there is a CP at time } t, \\
  r_t + 1, & \text{there is no CP at time } t, \\
  \end{cases}
\end{equation}
that is, the run length $r_t$ counts the number of time steps since the last CP.  The objective of the BOCPD method is therefore to perform the recursive estimation of $p(r_t|\mathbf{X}_{1:t})$, i.e., the posterior of the run length given the observations. From this posterior distribution, it is possible to estimate the run length, and as a consequence the CPs, using for instance the maximum-a-posteriori (MAP) criterion. However, due to the recursive nature of the $p(r_t|\mathbf{X}_{1:t})$ computation, the algorithm may be unable to accumulate enough probability mass on low values of $r_t$ when the time between CPs becomes on the order of magnitude (or less) of the number of the model parameters in $\bm{\theta}_{1:t}$. This makes almost impossible to detect CPs in high-dimensional observations. A good perspective is that it would be difficult to obtain reliable estimates of the (large number of) unknown parameters rapidly enough, as observations $\mathbf{X}_{1:t}$ come in. As a consequence, $p(r_t|\mathbf{X}_{1:t})$ is very low for any small value of $r_t$.

To overcome this limitation, we further assume that even if the data are high-dimensional, they belong to a low-dimensional manifold, which admits the following latent variable model
\begin{equation}
p(\mathbf{x}_{t}|\bm{\theta}_t) = \int p(\mathbf{x}_{t}|\mathbf{z}_t) p(\mathbf{z}_{t}|\bm{\theta}_t)d\mathbf{z}_{t},
\end{equation}
where $\mathbf{z}_t$ is the vector of latent variables associated with the observation $\mathbf{x}_{t}$. Now, the parameters that change between CPs parametrize the distribution of the latent variables $p(\mathbf{z}_t | \bm{\theta}_t)$, and since these variables have a dimensionality smaller than that of $\mathbf{x}_t$, the size of $\bm{\theta}_t$ is also reduced.

For the time being, we consider that $\z_t$ may be either continuous or discrete, and $p(\mathbf{x}_{t}|\mathbf{z}_{t})$ is fixed and known a priori\footnote{In Section \ref{sec:circadian} we relax this restriction and use a conditional distribution parametrized by some \emph{unknown} parameters $\bm{\phi}_t$.}. Even under this hierarchical model, we are still interested in the posterior $p(r_t|\mathbf{X}_{1:t})$, and we therefore need to extend the recursive estimation of \citet{adams2007bayesian} to account for the sequence of latent variable vectors, $\mathbf{Z}_{1:t}$. To compute the posterior distribution, we shall start with the joint distribution $p(r_t,\mathbf{Z}_{1:t},\mathbf{X}_{1:t},\bm{\theta}_{t})$. From this one, $p(r_t|\mathbf{X}_{1:t})$ may be directly obtained as
\begin{equation}
\label{eq:fraction}
p(r_t|\mathbf{X}_{1:t}) = \frac{p(r_t,\mathbf{X}_{1:t})}{\displaystyle \sum_{r_t} p(r_t,\mathbf{X}_{1:t})},
\end{equation}
where
\begin{equation}
\label{eq:marginal}
p(r_t,\mathbf{X}_{1:t}) = \iint p(r_t,\mathbf{Z}_{1:t},\mathbf{X}_{1:t},\bm{\theta}_{t}) \ d\bm{\theta}_{t} d\mathbf{Z}_{1:t}.
\end{equation}

Computing the posterior distribution in \eqref{eq:fraction}  can be quite involved, mainly due to the integrals in \eqref{eq:marginal}, and it is not trivial to achieve a recursive expression directly from them. Thus, to maintain a feasible recursitivity, our first step is to factorize the joint distribution in \eqref{eq:fraction} as
\begin{equation}
\label{eq:decomposition}
p(r_t,\mathbf{Z}_{1:t},\mathbf{X}_{1:t},\bm{\theta}_{t}) = p(\mathbf{X}_{1:t}|\mathbf{Z}_{1:t})p(r_t,\mathbf{Z}_{1:t},\bm{\theta}_{t}),
\end{equation}
where we have exploited that $p(\mathbf{x}_{t}|\mathbf{z}_{t})$ is assumed fixed and known. Now, we continue by marginalizing out the parameters of the factorized joint distribution in \eqref{eq:decomposition}  as
\begin{align}
p(r_t,\mathbf{Z}_{1:t},\mathbf{X}_{1:t}) &= \int p(r_t,\mathbf{Z}_{1:t},\mathbf{X}_{1:t},\bm{\theta}_{t}) \ d\bm{\theta}_{t} = \int p(\mathbf{X}_{1:t}|\mathbf{Z}_{1:t})p(r_t,\mathbf{Z}_{1:t},\bm{\theta}_{t}) \ d\bm{\theta}_{t} \nonumber \\ &= p(\mathbf{X}_{1:t}|\mathbf{Z}_{1:t}) p(r_t,\mathbf{Z}_{1:t}), \label{eq:decomposition_marginal}
\end{align}
where the marginalization of $\bm{\theta}_{t}$ is now independent of $\mathbf{X}_{1:t}$ and takes the form
\begin{equation}
\label{eq:recursive2}
p(r_t,\mathbf{Z}_{1:t}) = \int  p(r_t,\mathbf{Z}_{1:t},\bm{\theta}_{t}) \ d\bm{\theta}_{t}.
\end{equation}
Note that the recursive nature of the algorithm (see Appendix A) is now a direct consequence of the factorization in \eqref{eq:decomposition}, leading to
\begin{equation}
\label{eq:recursive}
p(r_t,\mathbf{Z}_{1:t}, \bm{\theta}_{t}) = \sum_{r_{t-1}}p(r_t|r_{t-1}) p(\mathbf{z}_t|\bm{\theta}_{t}) p(\bm{\theta}_{t}|r_{t-1},\mathbf{Z}^{(r)}_{1:t-1}) p(r_{t-1},\mathbf{Z}_{1:t-1}),
\end{equation}
where $\mathbf{Z}^{(r)}_{1:t-1}$ denotes the subset of latent variables for a given partition \citep{adams2007bayesian}. Importantly, the last term in the right-hand side of \eqref{eq:recursive} can be obtained, as in \eqref{eq:recursive2},  by marginalizing over the previous likelihood parameters:
\begin{equation}
p(r_{t-1},\mathbf{Z}_{1:t-1}) = \int p(r_{t-1},\mathbf{Z}_{1:t-1}, \bm{\theta}_{t-1})d\bm{\theta}_{t-1}.
\end{equation}
The prior of $r_t$ conditioned to $r_{t-1}$, also known as the change-point prior,  is given by
\begin{equation}
p(r_t|r_{t-1}) = \begin{cases}
H(r_{t-1} + 1), &  r_t = 0, \\
1- H(r_{t-1} + 1), & r_t = r_{t-1} + 1,
\end{cases}
\end{equation}
where $H(\tau)$ is the \textit{hazard} function, which we assume to be constant with a given time-scale hyperparameter $\tau$. Finally, $p(\bm{\theta}_{t}|r_{t-1},\mathbf{Z}^{(r)}_{1:t-1})$ is the posterior of the parameters at time $t$ given the previous run length and the corresponding partition of the latent variables $\mathbf{Z}^{(r)}_{1:t-1}$.

Plugging now \eqref{eq:recursive} into \eqref{eq:recursive2}, the probability $p(r_t,\mathbf{Z}_{1:t})$ becomes
\begin{equation}
p(r_t,\mathbf{Z}_{1:t}) =  \sum_{r_{t-1}}p(r_t|r_{t-1}) \pi^{(r)}_t p(r_{t-1},\mathbf{Z}_{1:t-1}),
\end{equation}
where the predictive posterior \citep{adams2007bayesian,turner2009adaptive} is given by
\begin{equation}
\label{eq:predictive}
\pi^{(r)}_t = p(\mathbf{z}_t|r_{t-1},\mathbf{Z}^{(r)}_{1:t-1}) = \int p(\mathbf{z}_t|\bm{\theta}_{t})p(\bm{\theta}_{t}|r_{t-1},\mathbf{Z}^{(r)}_{1:t-1})d\bm{\theta}_t.
\end{equation}

So far, we have obtained a recursive expression, but we still face the first problem due to the marginalization of $\bm{\theta}_t$. Concretely, the exact evaluation of $\pi^{(r)}_t$ is often intractable for non-Gaussian likelihoods and may require numerical solutions or even approximate inference \citep{turner2013online}. Actually, the computation of $\pi^{(r)}_t$ is a well-known issue for every extension of the method in \citet{adams2007bayesian}, where conjugate-exponential models were considered and it had therefore a closed-form expression. We address this issue later in Section \ref{sec:inference}. The second problem is the marginalization over the sequence of latent variables $\mathbf{Z}_{1:t}$ in \eqref{eq:marginal}. To overcome this problem, we propose to use a simple latent class model. The latent variables are given by a set of discrete values $z_t \in \{1,2, \dots, K\}$, with $K$ being the number of classes. Moreover, this choice is also motivated by the fact that this latent variable results in a small number of unknown parameters, leading to better performance of the CPD algorithm, as we have pointed out before. Also, choosing a latent class model is interesting for the interpretability point of view, as we will see in Section \ref{sec:experiments}. To sum up, each observable object $\mathbf{x}_t$ has now a single discrete representation $z_t$, which indicates the class it belongs to. Thus, we need to compute $p(r_t,\mathbf{z}_{1:t},\mathbf{X}_{1:t},\bm{\theta}_{t})$, where $\mathbf{z}_{1:t} = [z_1, \ldots, z_t]^\top$ is the \emph{univariate} sequence of latent class indicators.

\begin{figure}[t!]
	\centering
	\includegraphics[width=6cm]{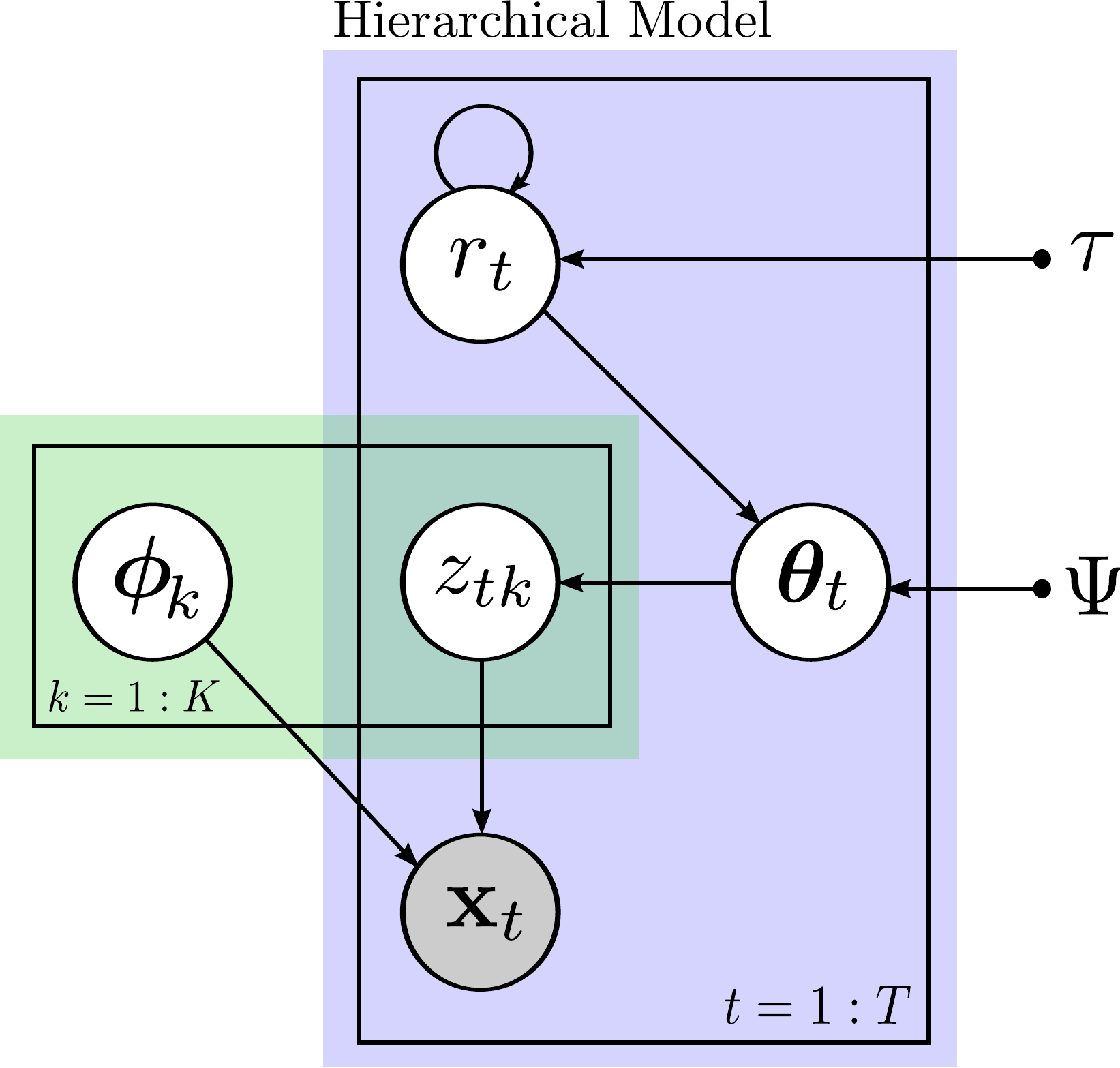}
	\vspace{.1in}
	\caption{Graphical representation of the hierarchical circadian model, where grey nodes represent observed variables, whereas white nodes correspond to hidden variables. The recursive conditioning of the run length $r_t$ with respect to its previous value is denoted by a self-connected variable. The latent variable $z_{tk}$ in the green plate is denoted using the one-hot-encoding convention. The fixed variable $\Psi$ denotes the set of hyperparameters involved in the prior of $\bm{\theta}_t$. Change-point detection performs exclusively on the blue region, while the green one captures the temporal structure of observations $\mathbf{x}_t$ as well as takes heterogeneity and complexity out of the hierarchical model (See Section \ref{sec:circadian}).} 
	\label{fig:graphicalmodel}
	\vspace{.1in}
\end{figure}

Based on this selection for the latent variable model, \eqref{eq:marginal} becomes
\begin{equation}
\label{eq:marginalizing}
p(r_t,\mathbf{X}_{1:t}) = \sum_{\mathbf{z}_{1:t}} p(r_t,\mathbf{z}_{1:t},\mathbf{X}_{1:t}) = \sum_{r_{t-1}}p(r_t|r_{t-1})\sum_{z_{t}}p(\mathbf{x}_t|z_t)\sum_{\mathbf{z}_{1:t-1}}\psi^{(r)}_{t},
\end{equation}
where 
\begin{equation}
  \psi^{(r)}_t = p(z_t|r_{t-1}, \mathbf{z}_{1:t-1}^{(r)})p(r_{t-1}, \mathbf{z}_{1:t-1}) p(\mathbf{X}_{1:{t-1}}| \mathbf{z}_{1:t-1}),
\end{equation}
and we have factored
\begin{equation}
  p(\mathbf{X}_{1:{t}}| \mathbf{z}_{1:t}) = p(\mathbf{x}_{t}| z_{t}) p(\mathbf{X}_{1:{t-1}}| \mathbf{z}_{1:t-1}).
\end{equation}
The values $\psi^{(r)}_t$ represent a function parametrized at each time step $t$ by all previous latent classes $\mathbf{z}_{1:t-1}$ for a given run length $r_{t-1}$. Due to the recursive dependence in \eqref{eq:marginalizing}, marginalizing this whole function results in a combinatorial problem that can be computationally challenging. Concretely, for sequences of length $T$ and with $K$ classes, the evaluation of $p(r_t,\mathbf{X}_{1:t})$ requires $\mathcal{O}(K^T)$ operations. Even if \eqref{eq:marginalizing} provides certain advantages with respect to \eqref{eq:marginal}, it may not be possible to compute it for long observation periods, $T$, and/or a large number of classes, $K$. For these cases, we present two simplified approaches in the following sections. In Figure \ref{fig:graphicalmodel}, we depict the proposed hierarchical model. Concretely, it shows that the run length $r_t$, which has a density parametrized by $\tau$, controls the generation of new parameters $\bm{\theta}_t$ for the latent variable density (i.e., a CP), modulating therefore the probability of each class $z_{tk}$. We also denote $\Psi$ as the subset of hyperparameters involved in the prior distribution over $\bm{\theta}_t$. The vector of parameters $\boldsymbol{\phi}_k$, from a predefined set $\{\boldsymbol{\phi}_k\}_{k = 1}^{K}$, parametrize the likelihood of the observations $\mathbf{x}_t$ within the $k$th class indicator variable $z_{tk}$. This will be explained in more detail in Section \ref{sec:circadian}.

\subsection{Simplified Hierarchical Models}
\label{sec:simplified}

We have shown how the computational complexity of the hierarchical model grows exponentially as the signal length, $T$, and the number of classes, $K$, increase, which may become prohibitive very quickly. The main factor of this computationally complexity is the marginalization of $\psi^{(r)}_t$, since it requires summing over all combinations of the latent classes $\mathbf{z}_{1:t-1}$. To avoid this marginalization, we simplify the previous model by considering two alternative strategies. When dealing with latent variables, if collapsing the entire latent domain is too costly, we propose to \emph{observe} it. That is, we can explicitly observe the values taken by the class indicators $z_t$, which before were hidden. This sort of observation model has been previously explored by \citet{nazabal2016human} for the combination of classifiers and it offers several ways to overcome computational costly operations.
 
 \subsubsection{Full Posterior Observation (FPO)} 
 
 As a first approximation to the hierarchical model, we can use the posterior probability of the latent variables, $p(z_{t}|\mathbf{x}_{t})$, as the inputs to the change-point detector. Thus, assuming that this posterior distribution has been previously inferred (see Section \ref{sec:circadian}), we may take these probabilities as multivariate random variables, similarly to \citet{nazabal2016human}. That is, the new observable inputs are $\tilde{\mathbf{z}}_{t} = p(z_t|\mathbf{x}_t)$, with $\tilde{\mathbf{z}}_{t} \in \mathcal{S}^K$ since $\sum_{k=1}^{K}\tilde{z}_{kt} = 1$. Here, $\mathcal{S}^K$ denotes the $K$-dimensional simplex.
 
\begin{figure}[t]
	\centering
	\includegraphics[width=6.5cm]{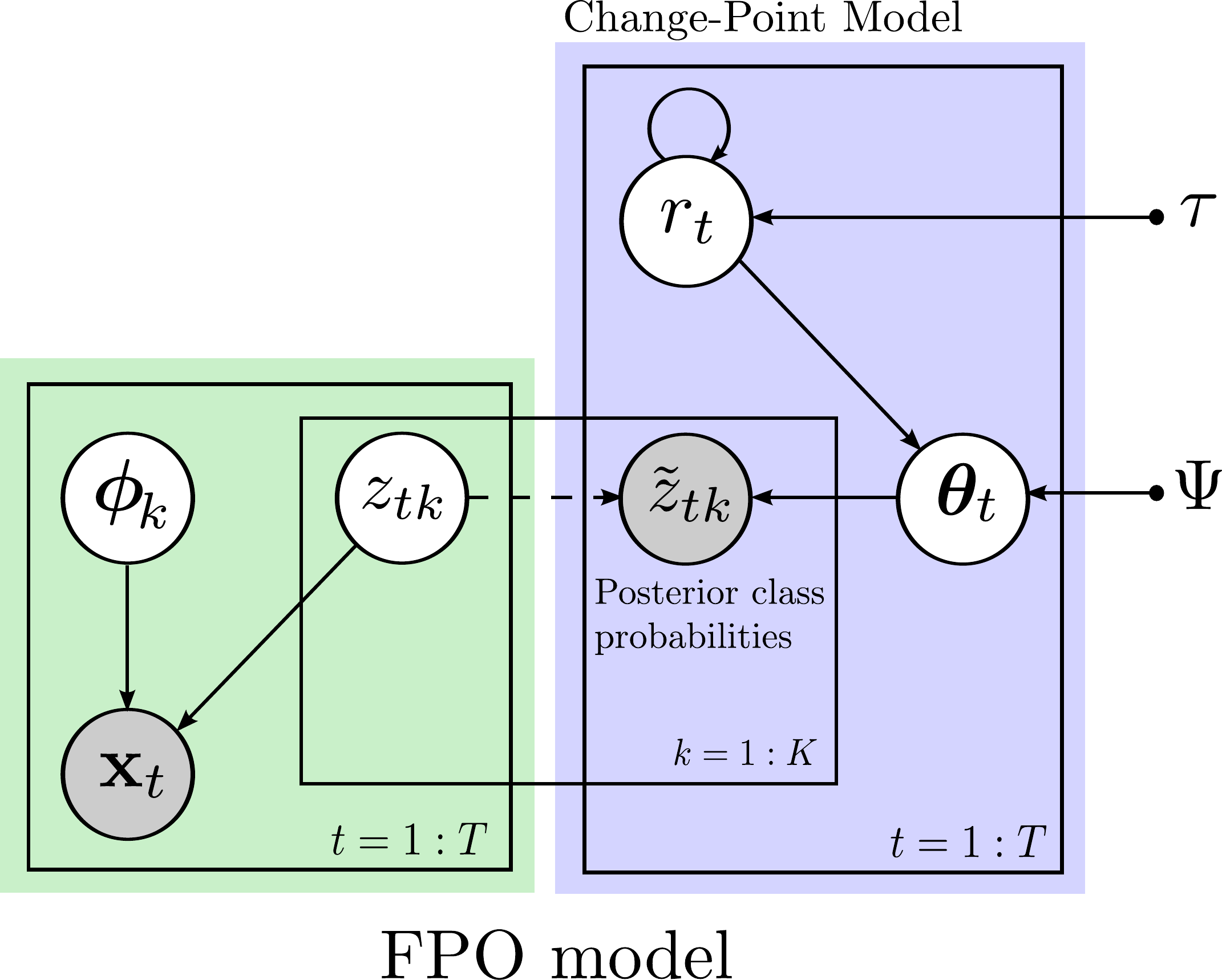}
	\hspace{1cm}
	\includegraphics[width=6.5cm]{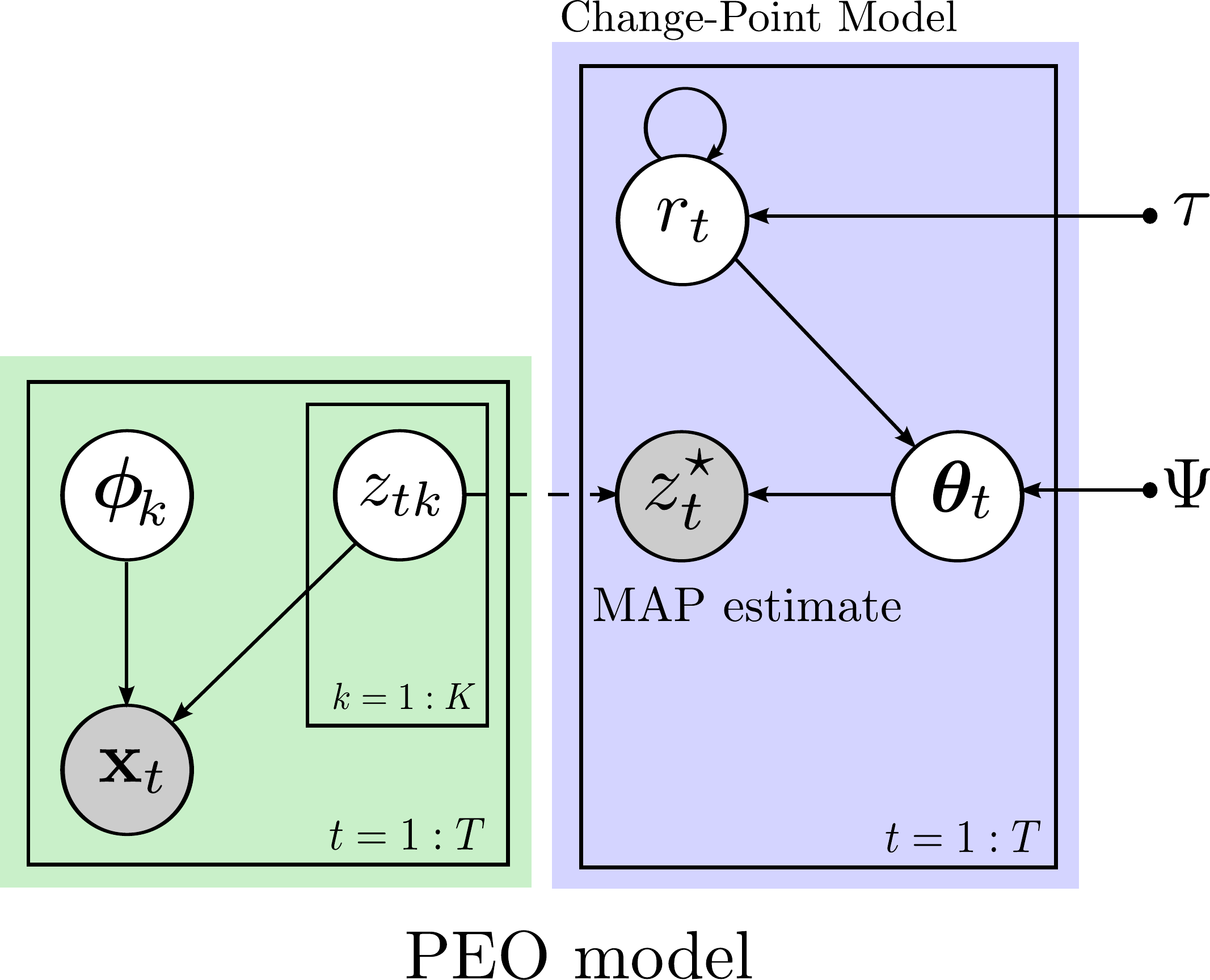}
	\vspace{.1in}
	\caption{Graphical representation of the simplified hierarchical models: Full Posterior Observation (FPO) and Point Estimate Observation (PEO). Shaded nodes correspond to observed variables and the dashed line denotes how the posterior probabilities of $z_t$ are observed by the change-point model (blue). In the left figure, the variable $\tilde{z}_{tk}$ is the $kth$ component of the vector $\tilde{\z}_t$ of probabilities for each class at time $t$, whereas in the right one, $z^{\star}_t$ is a point estimate which takes the $z_{tk}$ value with highest probability. In both cases, $\Psi$ denotes the fixed hyperparameters from the taken priors on the corresponding $\bm{\theta}_t$, similarly to Figure \ref{fig:graphicalmodel}.} 
	\label{fig:graphicalmodel_post}
\end{figure}

The main advantage of this approach is that even if the true classes $z_t$ are unknown, we can detect CPs directly from the sequence of posterior probabilities, which avoids the marginalization of the entire latent sequence. In particular, we may use the algorithm of \citet{adams2007bayesian}, but with a different likelihood model that does not allow for closed-form inference. Then, in Section \ref{sec:inference} we will present a sampling method to compute the posterior predictive in \eqref{eq:predictive}. Moreover, in Section \ref{sec:inference} we also show the good performance of this approximation. The graphical representation of this \textit{full posterior observation} (FPO) model can be seen in Figure \ref{fig:graphicalmodel_post}. Note that, contrary to the model in Figure \ref{fig:graphicalmodel}, here we partially observe the latent variables through $\tilde{\mathbf{z}}_{t}$, and that the change-point model is decoupled from the latent variable model.

\subsubsection{Point Estimate Observation (PEO)}

There may exist scenarios where the above approach is still too computationally demanding due to the sampling procedure. In these cases, which as we will see typically correspond to scenarios with a large number of classes $K$, an even simpler alternative may be considered if we are able to obtain good point estimates, $z^{\star}_t$, of the class indicators, $z_t$. Hence, instead of marginalizing $\mathbf{z}_{1:t}$, we can directly plug in the value of such estimates. To adapt the hierarchical model to these new observed latent variables, we take the set of point estimates $\mathbf{z}^{\star}_{1:t}$ as the input sequence to the detector. Note that we further assume the likelihood to be $p(z^{\star}_t|\bm{\theta})$, that is, we are directly modeling the changes on the sequence of point-estimates. In particular, we compute the point estimates using the maximum-a-posteriori criterion. That is, $z^{\star}_t = \arg\max_{z_t} p(z_t|\mathbf{x}_t)$, for which we need to previously infer the posterior distribution $p(z_t|\mathbf{x}_t)$. In the remaining lines of this section, we assume that it is given, and show, later in Section \ref{sec:circadian}, how to obtain it. The graphical model of this \textit{point-estimate observation} (PEO) approximation is shown in Figure \ref{fig:graphicalmodel_post}.

\subsection{Inference}
\label{sec:inference}

In this section, we describe the inference procedures for the two simplified hierarchical cases: the FPO and PEO models. For the former, it is only possible to do approximate inference, whereas for the latter, exact inference can be achieved.

\subsubsection{Approximate Inference for FPO Model}

The likelihood function for this model is $p(\tilde{\mathbf{z}}_t|\bm{\theta})$, which we take as a Dirichlet distribution with natural parameter $\bm{\theta}$. Thus, our objective is to compute $p(r_t|\tilde{\mathbf{z}}_{1:t})$, for which the hierarchical model described before must be reformulated. Moreover, the recursiveness in \eqref{eq:recursive} needs to be updated to accept the new sequence of observations $\tilde{\mathbf{z}}_{1:t}$. This is equivalent to using the BOCPD algorithm with inputs $\tilde{\mathbf{z}}_{t} = p(z_t|\mathbf{x}_t)$, but as we will see later, conjugacy is unavailable in this case. To simplify the inference process, we decompose the natural parameter of the Dirichlet distribution as $\bm{\theta}= \eta \bm{\lambda}$, with inverse variance $\eta \in \mathbb{R}_{+}$ and mean $\bm{\lambda} \in \mathcal{S}^K$. This decomposition allows us to choose a Gamma prior for $\eta$ and a Dirichlet one with single natural parameter $\bm{\beta}$ for the mean $\bm{\lambda}$. Under this construction, we may rewrite the probability model as 
\begin{equation}
\tilde{\mathbf{z}}_t\sim\text{Dir}(\eta,\bm{\lambda}),
\end{equation} 
where 
\begin{align}
\eta &\sim \text{Ga}(\kappa,\nu), & \bm{\lambda} &\sim \text{Dir}(\bm{\beta}),
\end{align} 
with $\bm{\beta}\in\mathcal{S}^K$ and $\kappa, \nu\in\mathbb{R}_+$. \\

We now want to obtain $p(r_t|\tilde{\mathbf{z}}_{1:t})$ in a similar way as in \eqref{eq:fraction}. For that, it is necessary to compute the predictive integral $\pi^{(r)}_t$, which is given by an expression equivalent to \eqref{eq:predictive}. To compute this integral, we need $p(\bm{\theta}_{t}|r_{t-1},\tilde{\mathbf{Z}}^{(r)}_{1:t-1})$, which reads as
\begin{multline}
p(\bm{\theta}_{t}|r_{t-1},\tilde{\mathbf{Z}}^{(r)}_{1:t-1}) = p(\eta_t,\bm{\lambda}_t|r_{t-1},\tilde{\mathbf{Z}}^{(r)}_{1:t-1}) \propto
p(\eta_t)p(\bm{\lambda}_t) \prod_{\tau = 1}^{r_{t-1}} p(\tilde{\mathbf{z}}_{\tau}|\eta_t,\bm{\lambda}_t) \\
= \text{Ga}(\eta_t|\kappa,\nu)\text{Dir}(\bm{\lambda}_t|\bm{\beta}) \prod_{\tau = 1}^{r_{t-1}} \text{Dir}(\tilde{\mathbf{z}}_{\tau}|\eta_t,\bm{\lambda}_t),
\end{multline}
where we have used both prior distributions defined above and considered the hyperparameters $\kappa, \nu,$ and $\bm{\beta}$ fixed. Therefore, the desired predictive integral becomes
\begin{equation}
\pi^{(r)}_t = p(\tilde{\mathbf{z}}_{t}|r_{t-1}, \tilde{\mathbf{Z}}^{(r)}_{1:t-1}) = 
\int p(\tilde{\mathbf{z}}_{t}|\eta_t,\bm{\lambda}_t) p(\eta_t,\bm{\lambda}_t|r_{t-1},\tilde{\mathbf{Z}}^{(r)}_{1:t-1}) d\eta_t d\bm{\lambda}_t.
\end{equation}
However, as we already pointed out, this integral is analytically intractable. Instead, we propose to solve it via Markov chain Monte Carlo (MCMC) methods as follows
\begin{equation}
\pi^{(r)}_t \approx \frac{1}{S} \sum_{s = 1}^{S} p(\tilde{\mathbf{z}}_{t}|\eta^{s},\bm{\lambda}^{s}),
\end{equation}
where $\eta^{s}$ and $\bm{\lambda}^{s}$ are the $s$th samples of $\eta$ and $\bm{\lambda}$, with $S$ being the total number of samples. In particular, we use a Gibbs sampler to draw realizations from $p(\eta_t,\bm{\lambda}_t|r_{t-1},\tilde{\mathbf{Z}}^{(r)}_{1:t-1})$. The equations for the conditional probabilities are
\begin{multline}
\label{eq:cond1} p(\eta_t| r_{t-1}, \tilde{\mathbf{Z}}^{(r)}_{1:t-1}, \bm{\lambda}_t^{s-1}) \propto
p(\eta_t)p(\tilde{\mathbf{Z}}^{(r)}_{1:t-1}|r_{t-1},\eta_t^{s-1}, \bm{\lambda}_t^{s-1}) \\ = \text{Ga}(\eta_t|\kappa,\nu) \prod_{\tau = 1}^{r_{t-1}} \text{Dir}(\tilde{\mathbf{z}}_{\tau}|\eta_t^{s-1},\bm{\lambda}_t^{s-1}), 
\end{multline}
and
\begin{multline}
\label{eq:cond2}  p(\bm{\lambda}_t|r_{t-1},\tilde{\mathbf{Z}}^{(r)}_{1:t-1}, \eta_t^{s-1}) \propto
p(\bm{\lambda}_t)p(\tilde{\mathbf{Z}}^{(r)}_{1:t-1}|r_{t-1},\eta_t^{s-1}, \bm{\lambda}_t^{s-1}) 
\\ = \text{Dir}(\bm{\lambda}_t|\bm{\beta}) \prod_{\tau = 1}^{r_{t-1}} \text{Dir}(\tilde{\mathbf{z}}_{\tau}|\eta_t^{s-1},\bm{\lambda}_t^{s-1}),
\end{multline}
where $\eta_t^{s-1}$ and $\bm{\lambda}_t^{s-1}$ are the realizations drawn in the previous iteration of the Gibbs sampler. Moreover, since there is no direct way to get samples from the conditional distributions in \eqref{eq:cond1} and \eqref{eq:cond2}, we propose to use the Gibbs-within-Metropolis Hastings sampler presented in \citet{martino2015independent,martino2018recycling}. The main idea is to use the random-walk (RW) Metropolis-Hastings (MH) algorithm \citep{martino2017metropolis} to generate samples from \eqref{eq:cond2}, as this distribution becomes very narrow when the number of latent classes, $K$, is large. On the other hand, samples from the conditional in \eqref{eq:cond1} may be obtained from a standard MH sampler with a Gamma proposal. 

The complete change-point detection algorithm for the FPO simplified model, including the Gibbs sampler, is summarized in Algorithm \ref{alg:posterior}. One final comment is that the inference process for this model, given a huge number of classes $K$, could still be high time demanding as a consequence of approximating $\pi^{(r)}_t$, with a large number of samples $S$, at each time step $t$.


\begin{algorithm}[!tb]
	\caption{Simplified hierarchical detection for the FPO Model}
	\label{alg:posterior}
	\begin{algorithmic}
		\STATE {\bfseries Input:} Observe $\mathbf{X}_{1:t} \rightarrow$ obtain $ \tilde{\mathbf{Z}}_{1:t} = p(\mathbf{z}_{1:t}|\mathbf{X}_{1:t})$
		\FOR{$\tilde{\mathbf{z}}_t$ {\bfseries in} $\tilde{\mathbf{Z}}_{1:t}$}
		\FOR{$r_t=1$ {\bfseries to} $t$}
		\STATE Evaluate $\pi^{(r)}_t = \frac{1}{S}\sum_{s=1}^{S}p(\tilde{\mathbf{z}}_t|\bm{\lambda}^{s},\eta^{s})$
		\ENDFOR
		\STATE Calculate growth probabilities: $p(r_t=r_t+1,\tilde{\mathbf{Z}}_{1:t})$ 
		\STATE Calculate change-point probabilities: $p(r_t=0,\tilde{\mathbf{Z}}_{1:t})$ 
		\STATE Calculate $p(\tilde{\mathbf{Z}}_{1:t}) = \sum_{r_t}p(r_t,\tilde{\mathbf{Z}}_{1:t})$
		\STATE Compute $p(r_t|\tilde{\mathbf{Z}}_{1:t})$ 
		\FOR{$r_t=1$ {\bfseries to} $t+1$}
		\STATE Sample $\eta^s \sim p(\eta|\bm{\lambda}^{s-1},r_t,\tilde{\mathbf{Z}}_{1:t})$ 
		\STATE Sample $\bm{\lambda}^s \sim p(\bm{\lambda}|\eta^{s},r_t,\tilde{\mathbf{Z}}_{1:t})$ using a RW MH algorithm
		\ENDFOR
		\ENDFOR
	\end{algorithmic}
\end{algorithm}

\subsubsection{Exact Inference for PEO Model}

To perform CPD for the second simplified model, we choose the likelihood function $p(z^{\star}_t|\bm{\theta})$ to be a categorical distribution with natural parameter $\bm{\pi}$. Similarly to the previous strategy, we place a Dirichlet prior on $\bm{\pi}$ but with a single hyperparameter $\bm{\gamma}$. Then, the generative model is
\begin{align}
z^{\star}_t &\sim\text{Cat}(\bm{\pi}), & \bm{\pi} &\sim\text{Dir}(\bm{\gamma}),
\end{align} 
where $\bm{\pi}\in\mathcal{S}^K$ and $\bm{\gamma}\in\mathbb{R}^K_+$, with $\mathbb{R}^K_+$ being the $K$-dimensional positive orthant. Interestingly, this choice for the priors allow to compute the predictive probabilities $\pi^{(r)}_t$ in closed form, which are given by
\begin{equation}
  \label{eq:predictive_categorical}
  \pi^{(r)}_t = p(z^{\star}_{t}|r_{t-1},\mathbf{z}^{\star (r)}_{1:t-1}) = \frac{\gamma^{(r)}_{k}}{\displaystyle \sum^{K}_{k'=1}\gamma^{(r)}_{k'}}, \forall r \in \{1,\cdots, t\},
\end{equation}
where $\gamma^{(r)}_{k}$ is the $k$th component of the parameters of the Dirichlet prior computed for the run length $r_t$. This closed-form expression, which is a direct consequence of the Dirichlet-Categorical conjugacy, provides a significant reduction in the computational complexity and results in a very simple method (see Algorithm \ref{alg:simplified}). The final step of this algorithm updates the parameters as
\begin{equation}
  \gamma^{(r)}_{k} \leftarrow \gamma^{(r)}_{k} + \mathbb{I}[z^{\star}_{t} = k],
\end{equation}
where $\mathbb{I}[\cdot]$ is the indicator function.

\begin{figure}[t!]
	\centering
	\hspace*{1.2cm}\includegraphics[width=9.25cm]{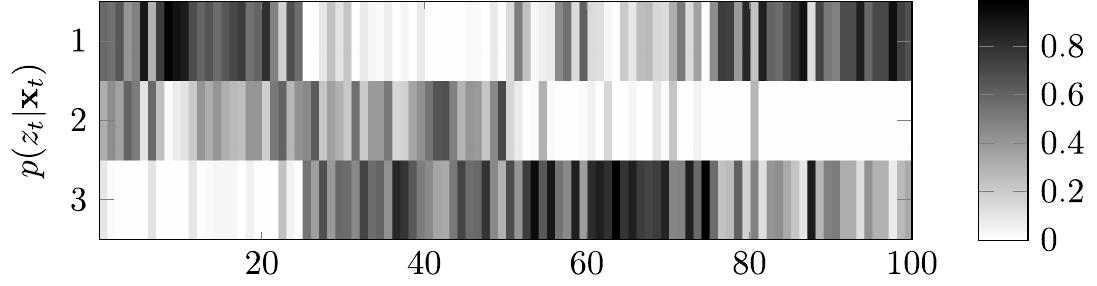}
	\hspace*{0.1cm}\includegraphics[width=7.9cm]{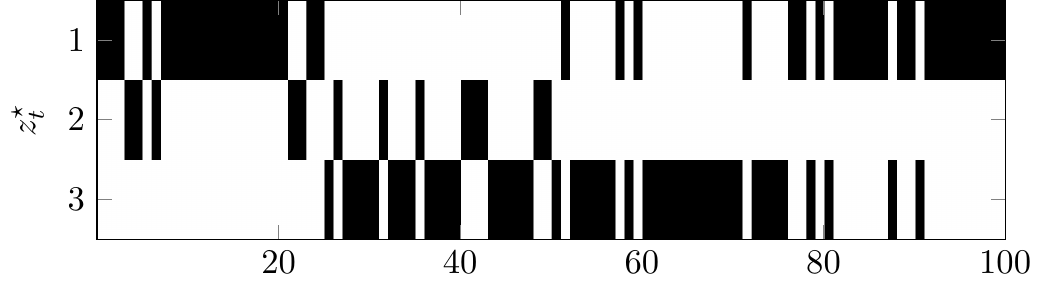}
	\includegraphics[width=7.9cm]{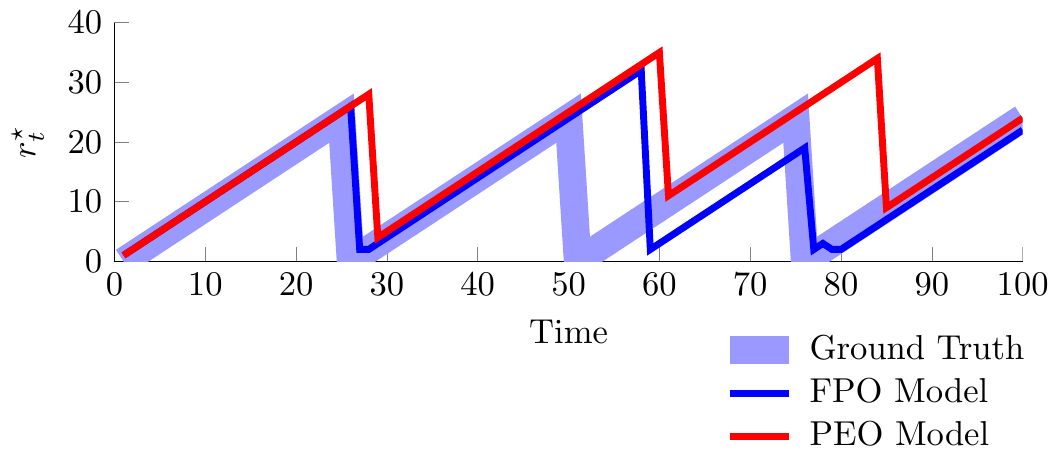}
	\caption{Comparison between change-point detection on FPO and PEO simplified models. \textbf{Top Row:} Sequence of posterior probability vectors $p(z_t|\mathbf{x}_t)$ where $K=3$ and $T=100$. Three change points are placed at $t=\{25,50,75\}$. \textbf{Middle Row:} Sequence of point estimates $z^{\star}_t = \arg\max_{z} p(z_t|\mathbf{x}_t)$ as $1$-of-$K$ encoding. \textbf{Bottom Row:} Ground truth of change points (blue), CP estimation $r^{\star}_t$ from FPO (blue) and from PEO model (red).} 
	\label{fig:simplified}
\end{figure}

It is important to mention that the PEO model is the most effective one in terms of computational cost, while at the same time the performance degradation is minimal, as shown in Figure \ref{fig:simplified}. Concretely, this figure shows the posterior probabilities of the latent variable (top row), the MAP estimates $\mathbf{z}^{\star}_{1:t}$ (middle row), and the outputs of the two simplified algorithms, as well as the ground truth. As can be seen in this example, the PEO algorithm detects the same change points, just with a slightly larger delay, but with a reduced computational complexity. This figure also illustrates what happens when there is a change point. For instance, there is a CP at position $t=50$, and observing the upper plot, we may see how the Dirichlet distribution has density mass concentrated on $z_t=\{2,3\}$ before the CP and, after the CP, the highest probability is located at $z_t=3$. 


\begin{algorithm}[t!]
	\caption{Simplified hierarchical detection for the PEO Model}
	\label{alg:simplified}
	\begin{algorithmic}
		\STATE {\bfseries Input:} Observe $\mathbf{X}_{1:t} \rightarrow$ obtain $ p(\mathbf{z}_{1:t}|\mathbf{X}_{1:t})$
		\FOR{$z^{\star}_{t}$ {\bfseries in} $\mathbf{z}^{\star}_{1:t} = \arg\max_z p(\mathbf{z}_{1:t}|\mathbf{X}_{1:t})$}
		\STATE Evaluate $\pi_t = p(z^{\star}_{t}|r_{t-1},\mathbf{z}^{\star (r)}_{1:t-1})$ using \eqref{eq:predictive_categorical}
		\STATE Calculate $p(r_t,\mathbf{z}^{\star}_{1:t})$ 
		\STATE Calculate $p(\mathbf{z}^{\star}_{1:t})$
		\STATE Compute $p(r_t|\mathbf{z}^{\star}_{1:t})$
		\STATE Update $p(\bm{\theta}_{t+1}|r_t,\mathbf{z}^{\star}_{1:t})$, i.e., update $\gamma^{(r)}_{k}$
		\ENDFOR
	\end{algorithmic}
\end{algorithm}

\subsection{Missing Temporal Data}

In this section, we assume that all missing observations follow the model completely at random (MCAR) \citep{rubin1976inference}, and we denote them by $\mathbf{x}^{m}_t$. On the other hand, we denote the samples that are fully observed as $\mathbf{x}^{o}_t$. Dividing the observations in two sets also translates to the sequence of latent variables, which are divided into lost observations $z^{m}_{t}$, where all components of $\mathbf{x}^{m}_{t}$ are missing, and observed variables $z^{o}_{t}$, which correspond to $\mathbf{x}^{o}_t$. The case where only some components of $\mathbf{x}_t$ are missing is addressed in Section \ref{sec:circadian} from a different perspective.

Following a fully Bayesian approach, we marginalize the missing latent variables out in the hierarchical model. The corresponding predictive probability is therefore reduced to 
\begin{equation}
\pi^{(r)}_t = \int p(\mathbf{z}^{m}_{t}|r_{t-1},\mathbf{Z}_{1:t-1}) d \mathbf{z}^{m}_{t} = 1.  
\end{equation}
The main advantage of this marginalization is that the detection algorithm is able to compute probabilities $p(r_t,|\mathbf{X}_{1:t-1})$, even if a certain $\mathbf{z}_t$ is missing. Moreover, the recursivity remains unaltered since the conditional change-point prior, $p(r_t|r_{t-1})$, is always evaluated sequentially. That is, the uncertainty is still propagated forwards as
\begin{equation}
\label{eq:missing_predictive}
	p(r_t,\mathbf{Z}_{1:t}) = \sum_{r_{t-1}} p(r_{t-1},\mathbf{Z}_{1:t-1}) p(r_t|r_{t-1}).
\end{equation}
Our approach for incomplete sequences presents good performance if the missing entries do not appear as long bursts. If that were the case, the lack of randomness induced by the missing data may ruin the estimation of change points, as $p(r_t,\mathbf{Z}_{1:t})$ in \eqref{eq:missing_predictive} would be computed from very old data. 

\begin{figure}[!t]
	\centering
	\hspace*{0.1cm}\includegraphics[width=9cm]{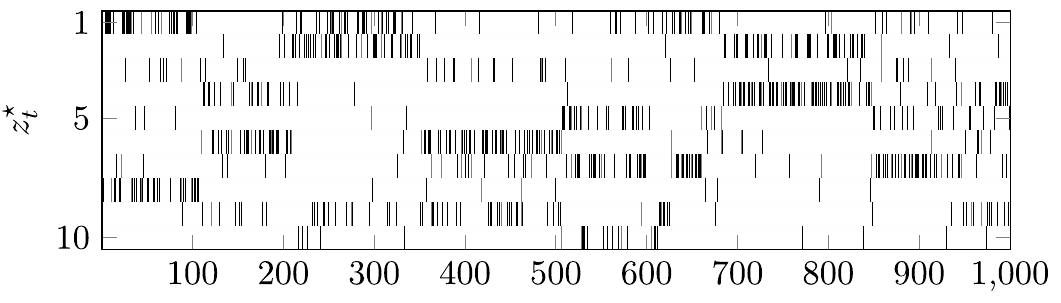}
	\includegraphics[width=9.1cm]{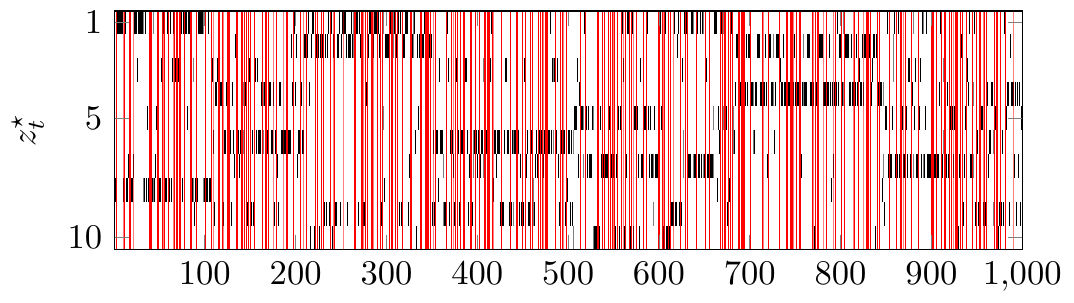}
	\includegraphics[width=9.2cm]{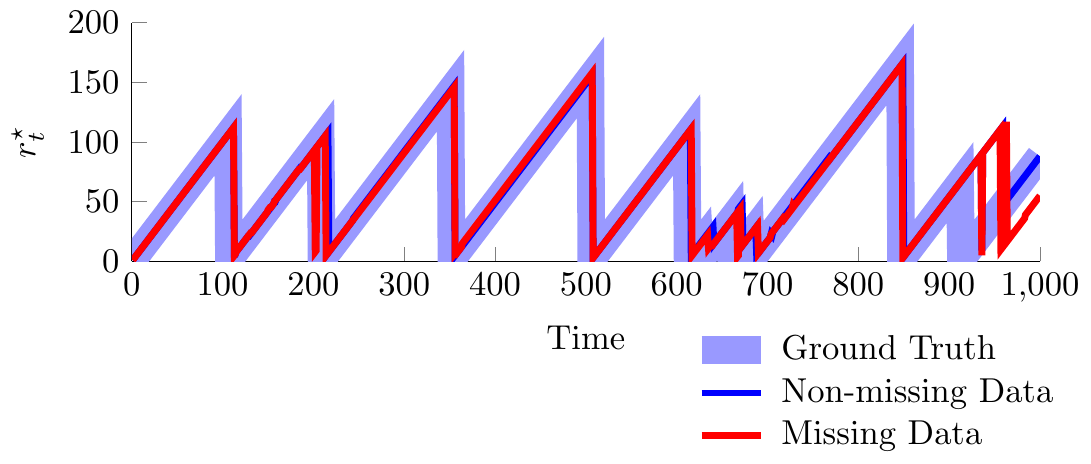}
	\caption{Change-point detection on the simplified hierarchical model with and without missing data. \textbf{Top Row:} Fully observed sequence $\mathbf{z}^{\star}_{1:t}$ of point-estimates ($1$-of-$K$ encoding) with $K=10$. \textbf{Middle Row:} Sequence $\mathbf{z}^{\star}_{1:t}$ with random  missing entries (red) ($25 \%$ rate).  \textbf{Bottom Row:} True $r_{1:t}$ (blue), MAP estimation of change-points from the complete sequence (black) and from the incomplete sequence (red).}
	\label{fig:missing}
\end{figure}

In Figure \ref{fig:missing}, we compare the PEO model for complete and incomplete data sequences. First, this figure shows the MAP estimates of $\mathbf{z}_{1:t}$ for the complete case (top row) and the incomplete one (middle row). Additionally, in the bottom row we plot the estimates of $r_t$ for both cases of missing and non-missing data, as well as the ground truth. It proves that the performance under missing data is almost identical to the case of non-missing data.

\section{Heterogeneous Circadian Models}
\label{sec:circadian}

We now study how to embed heterogeneous models into the hierarchical BOCPD technique presented before, as well as to handle periodic temporal structures. Motivated by the application to human behavior characterization, we consider sorts of data that possess a periodic temporal structure, with $24$-hours period, which is induced by the circadian rhythm. However, the ideas presented here are valid for any dataset with known periodicity. In our context, there exists significant work related to capturing such circadian rhythms at different scales in time series. For example, its application to gene expression modeling \citep{hensman2013hierarchical,durrande2016detecting} aims to identify the periodic dynamic cycles of cellular mechanisms in nature, a key point for the understanding of their hidden correlations. 

\begin{figure}[t!]
	\centering
	\includegraphics[width=12cm]{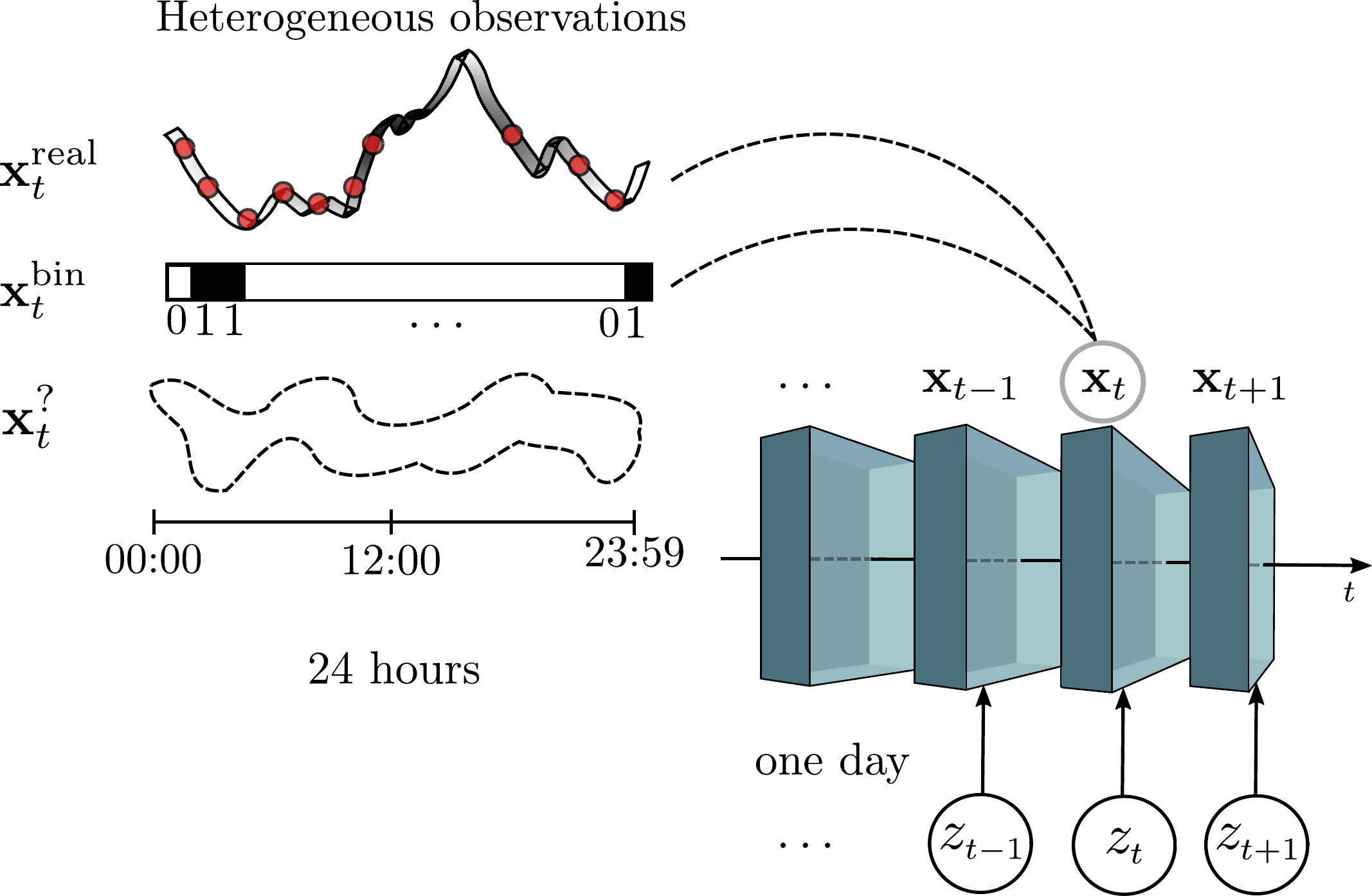}
	\caption{Schematic representation of the embedded circadian model. Each shaded box represents the observation at time $t$, $\x_t$. This observation is obtained by stacking the vectors that correspond to each data type, i.e., real, binary, etc. For instance, the vector composed by real numbers corresponds, in the application to human behavior characterization, to the traveled distance during every hour, yielding a $24$-dimensional vector. Moreover, associated to $\x_t$, there exists a single latent variable $z_t$ that is lower dimensional and we aim to discover.} 
	\label{fig:infography}
\end{figure}

To account for the periodic dependencies, we propose to arrange the data such that a single sample $\x_t$ at time $t$ stacks the observations of one period, as shown in Figure \ref{fig:infography}. Particularly, we build our heterogeneous observations by simply stacking dimensions from different statistical types on larger vectors $\x_t = [\x^{1}_t, \cdots, \x^{M}_t]^{\top}$, with $M$ being the number of heterogeneous data types. This essentially means that the sequence of observations $\mathbf{X}_{1:t}$ summarizes consecutive periods (i.e. one day) of different heterogeneous measurements. 

For the description of one day, $\mathbf{x}_t$, we propose a heterogeneous mixture model for the likelihood of the form
\begin{equation}
\label{eq:likeli_heter}
p(\x_{t}| z_{t},\{\bm{\phi}_k^1, \ldots, \bm{\phi}_k^M\}^{K}_{k=1}) = \prod^{K}_{k=1}\prod^{M}_{j=1}p(\mathbf{x}^j_{t}|\bm{\phi}^j_k)^{\mathbb{I}\{z_t = k\}},
\end{equation}
where the latent variables $z_t$ are indicators of which component is active. Concretely, it is composed by $K$ components, each one with its own likelihood function $p(\mathbf{x}^j_{t}|\bm{\phi}^j_k)$ for each data type, and with $\bm{\phi}^j_k$ denoting the parameters of the $j$th data type for the $k$th class. As can be seen in \eqref{eq:likeli_heter}, we have assumed that given the class $z_t$ and the parameters, the different types observations, $\x^{1}_t, \ldots, \x^{M}_t,$ are conditionally independent \citep{valera2017,nazabal2018}. This assumption simplifies the inference procedure, while at the same time it seems reasonable for the proposed application.

\subsection{Circadian Gaussian-Bernoulli Mixture Model}
\label{subsec:circadianGBMM}

Among all heterogeneous statistical data types that may be incorporated to the daily representation $\mathbf{x}_t$, we consider here two representative cases for the aforementioned application in Section \ref{sec:intro}.  We assume that each observation $\mathbf{x}_t$ is composed by binary and real values, that is, $\mathbf{x}_t = [(\mathbf{x}^{\text{real}}_t)^\top, (\mathbf{x}^{\text{bin}}_t)^\top]^\top$ where $\mathbf{x}^{\text{real}}_t \in \mathbb{R}^{D}$ and $\mathbf{x}^{\text{bin}}_t = [x_{t1}^{\text{bin}}, \ldots, x_{tD}^{\text{bin}}]^\top$,  with $x_{tj}^{\text{bin}} \in \{0,1\}$. Here, $D$ denotes the dimensionality of each kind of observation, and is given by a function of the period. For instance, if each component of $\mathbf{x}^{\text{real}}_t$ represents the traveled distance during each hour, then $D = 24$, which corresponds to one day, the period induced by the circadian rhythm.  Note also that we have assumed that all data types have the same dimensionality, but it is straightforward to consider different dimensionalities. For instance, we could consider $30$-minutes intervals for the binary variables and $1$-hour ones for the real vector.

Next, we want to propose a likelihood for each data type, which are necessary for \eqref{eq:likeli_heter}. Specifically, for the application that we have in mind in this paper, we choose Bernoulli and multivariate Gaussian distributions, yielding
\begin{equation}
p(\mathbf{x}_{t}|z_{t} = k,\{\bm{\phi}_k\}^{K}_{k=1}) = p(\mathbf{x}^{\text{bin}}_{t},\mathbf{x}^{\text{real}}_{t}|\bm{\phi}_k) =  \mathcal{N}(\mathbf{x}^{\text{real}}_{t}|\0,\mathbf{K}_k+\mathbf{D})\,\prod_{j=1}^{D}\text{Ber}(x^{\text{bin}}_{tj}|\mu_{kj}),
\end{equation}
where, for the time being, we denote $\bm{\phi}_k = \{\bm{\mu}_k, \bm{\gamma}_k\}$ as the set of likelihood parameters for each latent class value $k$. Here, $\bm{\mu}_k = [\mu_{k1}, \ldots, \mu_{kD}]^\top$, where each $\mu_{kj} \in [0,1]$ is the mean of the $j$th Bernoulli variable and $\bm{\gamma}_k$ are the parameters of the covariance matrices $\mathbf{D}$, that are a common diagonal matrices to all latent classes given by $\mathbf{D}=\text{diag}(\sigma_1^2, \ldots, \sigma_D^2)$. Covariance matrices $\mathbf{K}_k$ are positive-definite and corresponds to every class $k$. It is important to point out that the diagonal matrix $\D$ introduces some heteroscedasticity to the model since we are assuming that the noise may vary within time inputs.

To capture the periodic (circadian) feature of $\mathbf{x}^{\text{real}}_{t}$, we employ a periodic covariance function, similar to the works in \citet{durrande2016detecting} and \citet{solin2014explicit}. The covariance matrix $\mathbf{K}_k$ is therefore generated by a non-stationary periodic kernel, i.e., $[\mathbf{K}_k]_{t,t'} = g_k(t,t')$, where the time $t$ is assumed to be equally spaced (e.g., $t = 1,2,3,\ldots,D$). Further details about this function are provided in Section \ref{subsec:periodic}. Moreover, to obtain interesting insights, we propose that the distribution $\mathcal{N}(\mathbf{x}^{\text{real}}_{t}|\0,\mathbf{K}_k+\mathbf{D})$ is generated using a hierarchy similarly to \citet{hensman2013hierarchical,hensman2015fast}. Concretely, the mean $\mathbf{f}$ is drawn from a zero-mean Gaussian, which yields
\begin{align}
\mathbf{f} &\sim \mathcal{N}(\bm{0},\mathbf{K}_k), & \mathbf{x}^{\mathrm{real}}_t|\mathbf{f} &\sim \mathcal{N}(\mathbf{f},\mathbf{D}).
\end{align}
Note that the computation of $\mathbf{f}$ can be avoided by marginalizing it out, and therefore, both expressions would be reduced to $\mathbf{x}^{\mathrm{real}}_t \sim \mathcal{N}(\mathbf{0},\mathbf{K}_k +\mathbf{D})$.

Following the standard approach used in mixture models, we want to compute the complete likelihood, which includes the prior probability for each latent class indicator. By defining the prior distribution on class assignment $z_t$ as $p(z_t = k) = \pi_k$, the complete likelihood at time $t$ becomes
\begin{equation}
p(\mathbf{x}_{t}, z_{t} = k | \{\bm{\phi}_k\}^{K}_{k=1}) = p(z_t = k) p(\mathbf{x}_{t}| z_{t} = k,\bm{\phi}_k) = \pi_k p(\mathbf{x}_{t}|\bm{\phi}_k)\nonumber.
\end{equation}
Next, taking into account that observations are conditionally i.\,i.\,d., the complete joint log-likelihood becomes
\begin{multline}
\mathcal{L}_{\z, \bm{\phi}} = \log p(\mathbf{X}_{1:t},\mathbf{z}_{1:t}|\{\bm{\phi}_k,\pi_k\}^{K}_{k=1}) 
= \sum_{t=1}^{T}\sum_{k=1}^{K}\mathbb{I}\{z_t=k\} \log \pi_k \\ 
+ \sum_{t=1}^{T}\sum_{k=1}^{K}\mathbb{I}\{z_t=k\} \log p(\mathbf{x}^{\text{bin}}_t|\bm{\mu}_k) + \sum_{t=1}^{T}\sum_{k=1}^{K}\mathbb{I}\{z_t=k\} \log p(\mathbf{x}^{\text{real}}_t|\bm{\K}_k,\bm{\D}). \label{eq:loglik}
\end{multline}
Finally, a key point of this circadian model is that, if we needed to take observations at non-uniformly separated inputs $\mathbf{t}$, this approach would accept a different $\mathbf{t}$ for each $\mathbf{x}_t$ in a straightforward manner. 

\subsection{Periodic Non-stationary Covariance Functions}
\label{subsec:periodic}

The proposed model captures the circadian feature of data for each class $k$ through the temporal embedding and the periodic covariance functions $g_k(t,t')$, which must be non-stationary in our case. Notice that afternoon-evening should have a different correlation pattern than, for instance, nocturnal hours and early morning, which prevents the use of stationary approximations. One possible solution to build non-stationary kernels is using an input-dependent mapping $s_k(t)$, similarly to the one used in \citet{heinonen2016non}. Here, we have $g_{k}(t,t') = s_k(t) s_k(t') \tilde{g}_k(t-t')$, where $\tilde{g}_k(t-t')$ is a stationary periodic covariance function, associated to class $k$, that models the intrinsic temporal structure during a day. In this case, we take the periodic version of the exponential kernel \citep{mackay1998introduction}, which is given by 
\begin{equation}
\tilde{g}_k(t-t') = \sigma^2_{ak}\exp\Big(-\frac{2\sin^2(\pi(|t - t'|/D)}{\ell_k^2}\Big),
\end{equation}
and for $s_k(t)$, the hour-specific term, we use a squared Fourier series of order $C$, with $C \leq D$. This last constraint imposes a limit on the smoothness and avoids overfitting. Thus, $s_k(t)$ is
\begin{equation}
s_k(t) = \left(\frac{a_{k,0}}{2} + \sum_{c=1}^{C}\left[ a_{k,c}\cos\left(\frac{2\pi c}{D} t\right) + b_{k,c}\sin\left(\frac{2\pi c}{D} t\right) \right]\right)^2,
\label{eq:fourier}
\end{equation}
where $\mathbf{a}_k = [a_{k,0} \, \dots \, a_{k,C}]^\top$ and $\mathbf{b}_k = [b_{k,1} \, \dots \, b_{k,C}]^\top$ are Fourier coefficients that parametrize the covariance matrix of the $k$ class, $\K_k$, together with the parameters of the exponential kernel, $\sigma_{ak}$ and $l_k$.

It is also important to mention that the function $s_k(t)$ may introduce multimodality along time. Moreover, we are simultaneously modeling the correlation among daily hours, which is presumed to be non-stationary, and, at the same time, capturing the individual variance of data along different instants of the day. This is confirmed by the experimental results presented in Section \ref{sec:experiments}.

\subsection{EM Algorithm for Heterogeneous Circadian Models with Missing Data}

In this section, we present an inference technique for the model described above. In particular, given the model and the complete log-likelihood $\mathcal{L}_{\z, \bm{\phi}}$ in \eqref{eq:loglik}, we need to infer the assignments $z_t$ and the set of model parameters $\{\pi_k,\bm{\phi}_k\}^{K}_{k=1}$, where we now define as $\bm{\phi}_k = [\bm{\mu}_k^\top,  \mathbf{a}_k^\top, \mathbf{b}_k^\top, \sigma_{ak}, \ell_k  ,\bm{\sigma}^\top]^\top$, with $\bm{\sigma} = [\sigma_1, \ldots, \sigma_D]^\top$.

Obtaining maximum likelihood estimates of the Gaussian-Bernoulli mixture-model parameters in \eqref{eq:loglik} can be done through the expectation-maximization (EM) algorithm \citep{dempster1977maximum}. However, there is a slight difference in our model. While the expectation step is computed from the complete mixture distribution, in the maximization step, it is possible to factorize $\mathcal{L}_{\z, \bm{\phi}}$ and therefore to estimate the parameters of the Bernoulli and Gaussian distributions separately. This is a key point of the inference process, since as long as we introduce new likelihoods in the model, estimating the parameters of all likelihoods will be always a set of independent tasks. 

An important matter of using the EM algorithm here is that it allows us to handle missing components in any observation $\x_t$ \citep{ghahramani1994supervised}. That is, rather than dealing with sequences where we may find complete missing observations, sometimes we have to deal with just a few lost features, i.e., only a few components of $\x_t$ are missing, making the vector incomplete. The proposed technique estimates the probabilities $p(z_t|\mathbf{x}_t)$ even if $ \mathbf{x}_t$ is only partially observed. Hence, we are interested in adapting the mixture model for dealing with missing data following a similar approach to \citet{ghahramani1994supervised}, which computes expectations not only over latent variables but over missing values. 

Concretely, the algorithm iterates between the following steps
\begin{align}
\text{\textbf{E-step:}}\; & \text{compute  \hspace*{0.15cm}} \mathcal{Q}=\mathbb{E}_{\mathbf{z}_{1:t}, \mathbf{X}^m_{1:t}}[\mathcal{L}_{\z, \bm{\phi}}]\nonumber\\
\text{\textbf{M-step:}}\; & \text{maximize } \mathcal{Q}  \text{\hspace*{0.15cm} w.r.t.\  \hspace*{0.15cm}}  \{\mathbf{a}_k, \mathbf{b}_k, \sigma_{ak}, \ell_k\}^K_{k=1}, \text{and }  \bm{\sigma} \nonumber\\
& \text{maximize } \mathcal{Q}   \text{\hspace*{0.15cm} w.r.t.  \hspace*{0.15cm}} \{\bm{\mu}_k,\bm{\pi}_k\}^K_{k=1}\nonumber
\end{align}
where $\mathbf{X}^m_{1:t}$ denotes the matrix that collects the missing values of every observation. It is important to point out that there are no closed-form expressions for the maximization w.r.t. the parameters of the Gaussian distribution. Thus, we propose to use the conjugate gradient (CG) ascent method (see Appendix B and C). Finally, contrary to standard EM techniques, the proposed approach must combine closed-form estimators for the Bernoulli distribution parameters and numerical optimization for the hyperparameters of the periodic covariance function. Since the latter problem is highly non-convex, several initializations of the hyperparameters are required achieve a good (local) maximum. Some additional details about the initializations are presented in Appendix D.

\section{Experiments}
\label{sec:experiments}

In this section, we demonstrate that the proposed approach is capable of properly estimating change points in two different scenarios (one synthetic and one with real data) in the presence of heterogeneous and high-dimensional observations with strong periodic dependencies between samples. Concretely, for these examples, only the algorithm based on the PEO model is used since the computational complexities of the full hierarchical approach and the FPO model are prohibitive.

\subsection{Synthetic data}
\label{eq:synthetic}

In the first experiment, we want to validate the hierarchical change-point detector using synthetic data. Setting a categorical distribution as the likelihood distribution, that is, using the PEO model, we generate a discrete sequence $\mathbf{z}_{1:t}$ with  $T=500$ instances and $K=5$ classes (future $z_t$ assignments) where four change points were introduced. Note that this implies that the sequence is divided into five partitions. Then, the generative parameters $\bm{\pi}$ for each partition are generated by sampling $\bm{\pi} \sim \mathrm{Dir} \left(\alpha/5,\alpha/5,\dots,\alpha/5\right)$, and we fixed $\alpha=25$. 

The resulting sequence $\mathbf{z}_{1:t}$ corresponds to the latent classes in the corresponding hierarchical structure. Consequently, we generate pairs of multivariate Bernoulli-Gaussian samples where the covariance matrices are given by the non-stationary periodic kernel described in Section \ref{subsec:periodic} with a random selection of the hyperparameters. 

Note that, in our experiment, we only observe the given stream of binary and real observations, never the sequence of latent classes $\mathbf{z}_{1:t}$. Then, assuming the Gaussian-Bernoulli mixture model from Section \ref{subsec:circadianGBMM}, we infer the posterior probabilities over class assignments $p(z_t|\mathbf{x}^{\text{bin}}_t,\mathbf{x}^{\text{real}}_t)$ by applying the EM algorithm described in the previous Section. In the experiments, we have truncated the number of Fourier coefficients to $C=3$, while the data was generated with $C=2$. 

\begin{figure}[!ht]
	\centering
	\hspace*{1cm}$t=1$ \hspace*{1.4cm} $t=2$ \hspace*{0.5cm} $\cdots$ \hspace*{0.4cm} $t=T$\\ \vspace*{0.5cm}
	\hspace*{4.3mm}\includegraphics[width=7.5cm]{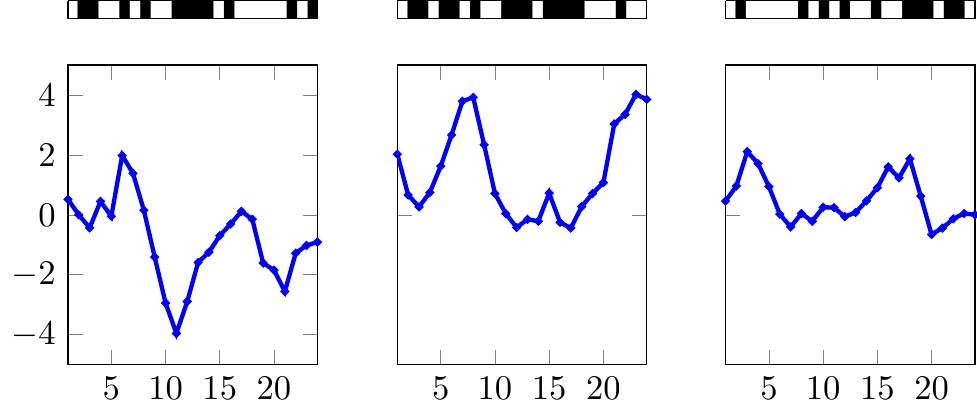}\\  \vspace*{0.25cm}
	\hspace*{0.3cm}\includegraphics[width=8.1cm]{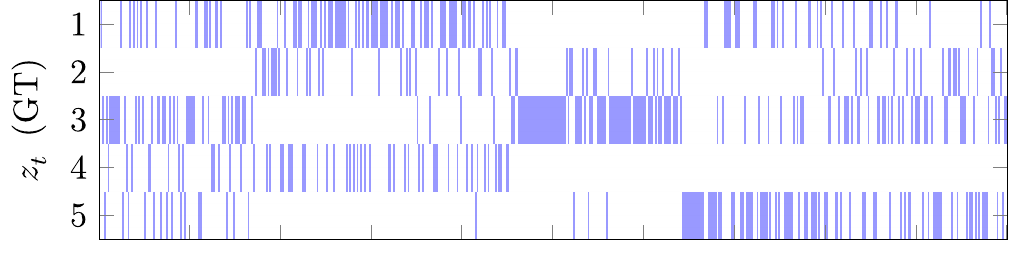}
	\hspace*{0.3cm}\includegraphics[width=8.1cm]{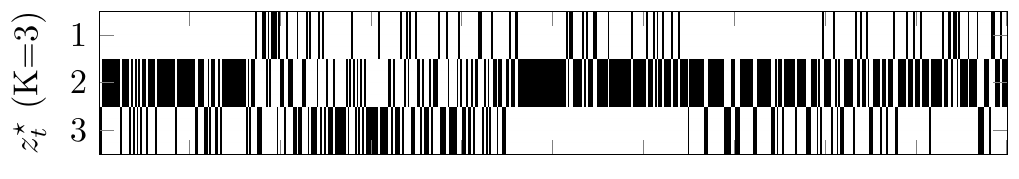}
	\hspace*{0.3cm}\includegraphics[width=8.1cm]{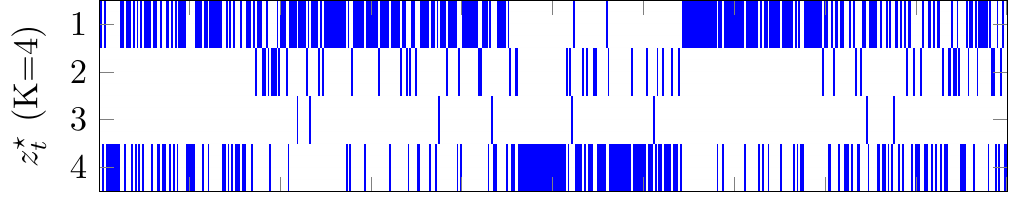}
	\hspace*{0.5cm}\includegraphics[width=8.3cm]{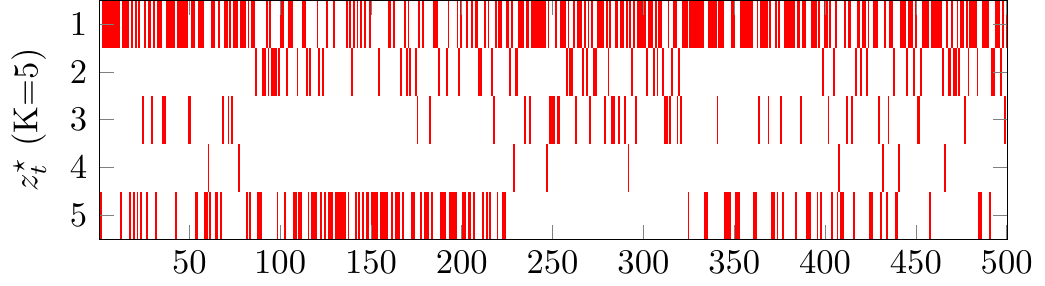}
	\includegraphics[width=8.1cm]{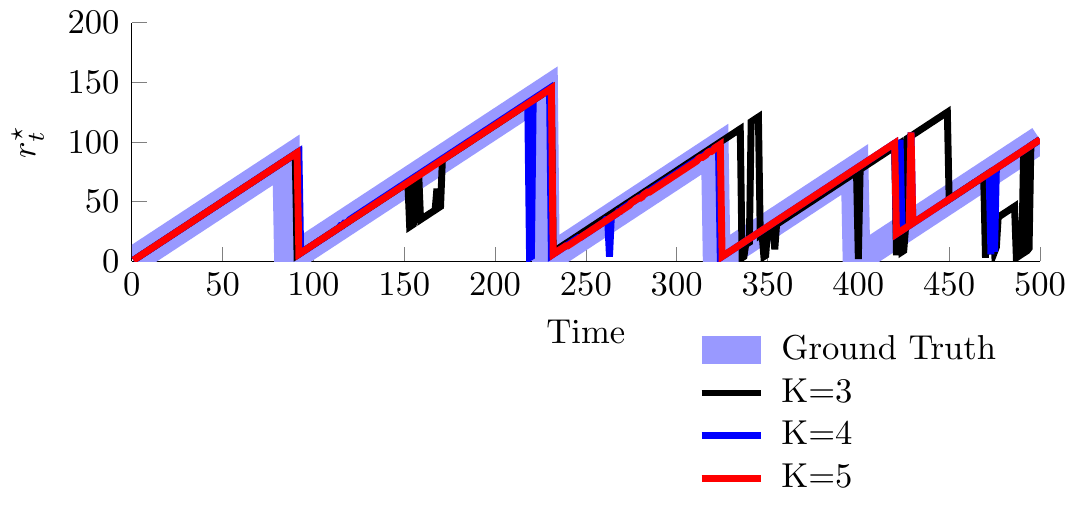}
	\caption{Results for the synthetic experiment. \textbf{First Row:} Each observation $\mathbf{x}_t$ consists of $\mathbf{x}^{\mathrm{real}}_t \in \mathbb{R}^D$ and $\mathbf{x}^{\mathrm{bin}}_t \in \{0,1\}^D$, with $D=24$. \textbf{Second Row:} Sequence of true latent variables. \textbf{Third to Fifth Rows:} Sequence of MAP estimates of the latent variables for $K = 3, 4,$ and $5$. \textbf{Last Row:} Ground Truth of CP segments and MAP-CPD traces of $r^\star_t$ for $K=3, 4,$ and $5$ latent classes.} 
	\label{fig:synthetic}
\end{figure}

From the sequence of posterior probability vectors $p(\mathbf{z}_{1:t}|\mathbf{X}_{1:t})$, the point-estimates $\mathbf{z}^{\star}_{1:t}$ were obtained  using the MAP estimator. Thus, we performed the hierarchical change-point detection of Algorithm \ref{alg:simplified}. We run the whole process for several $K$ classes in the Gaussian-Bernoulli mixture model and multiple initializations. Figure \ref{fig:synthetic} shows three examples of the heterogeneous observations for $t = 1, 2,$ and $T$. Moreover, it also shows the true latent variables, as well as the MAP estimates, $z_t^{\star}$, assuming that $K = 3, 4,$ and $5$. Finally, the last row of Figure \ref{fig:synthetic} plots the run length, where we can see the aforementioned CPs, as well as its MAP estimates for $K = 3, 4,$ and $5$. As we can see, the proposed method detects with a high accuracy the CPs, although, as the number of categories decreases, the detection presents a larger delay, mainly for the last two partitions. Our hypothesis is that as long as the circadian model gets a sufficiently large number of latent classes (i.e., it is more representative), the change-point detection can differ better between adjacent partitions. 

The interpretability of change-point detection diagrams is a key point of this paper (see Figures \ref{fig:simplified}, \ref{fig:missing},  and \ref{fig:synthetic}). Usually, we represent the run length with highest probability, that is, $r^\star_t = \arg \max p(r_t|\mathbf{X}_{1:t})$ at each time step $t$. The resulting diagram shows increasing run lengths while CPs are not detected. Note that ground truth plots in these figures always represent the exact time instant when changes occur. This is generally difficult to be detected without any sort of delay (i.e., a few samples from the post-change sequence must be observed to update the new distribution and then compare with the pre-change distribution). The usual case where this delay could be mostly reduced is given by the predictive probabilities $\pi^{(r)}_t$ in (\ref{eq:predictive}). That is, under extremely abrupt changing conditions, a new sample that does not follow the old distribution will give a very low probability (i.e., $\pi^{(r)}_t\approx 0$) and therefore will force almost immediately the run length to be zero, $r^\star_t = 0$. In most cases, this does not occur so we have to look for the lowest value taken by $r^\star_t$, which is usually a bit larger than zero. For example, in the last change point detected in Figure \ref{fig:synthetic} for $K = 5$, $r^\star_{420}=20$, which indicates that there was a change point $20$ time steps before, that is, at $t=400$.


\subsection{Anomalous Human Behavior}

In this experiment, we evaluate the proposed method on real data for medical applications. Since our work is highly motivated by the problem of change-point detection for personalized medicine, it is important to mention why detecting abnormal changes is a key milestone for the future of modern psychiatry. 

\subsubsection{Change-point Detection in Psychiatry}

Nearly all patients with mental health disorders such as schizophrenia or chronic depression are likely to suffer some sort of relapse in the next five years after the first episode. However, despite the fact that significant early indicators may appear before an abrupt relapse, when recognizable symptoms are already present, it is usually too late for any preventive treatment. 

The majority of these previous signs and triggering effects can be identified by continually monitoring changes in the patient's behavior. Here, we refer to  human behavior as the set of routines and patterns that are persistent during our daily life. In \citet{eagle2009eigenbehaviors}, it was demonstrated how these behavioral patterns are projected in the set of signals that we generate every day (i.e., mobility data, phone communications and logs of social interactions, etc.) and are mainly recognizable through the appearance of circadian features and periodicities.  Along these lines, the ubiquitous condition of smartphones and wearables, whose recording and memory capabilities are able to perform continuous monitoring of the aforementioned domains, opened a new window for analyzing all these sources of information in chronic patients without the need of direct medical interventions. Our primary goal would be to provide a robust solution to this paradigm by applying change-point detection on behavioral data obtained from monitored psychiatric patients.

\subsubsection{Mobility metrics}

 In this experiment, we used data that consists of location traces (latitude-longitude) recorded via the smartphone of a student of our lab during 275 consecutive days. It contains a bit more than 100,000 instances that correspond to the user's GPS coordinates every 3 minutes on average. We investigated how to obtain reliable metrics from these traces.  Concretely, we considered two types of metrics \citet{canzian2015trajectories}: 1) a real-valued signal of the log-distance travelled per hour and 2) daily binary vectors of presence or absence at home. 
 
The preprocessing step was different for each one of the metrics. In the first case, we collected all location traces separately for every hour and every day. Once this was done, we calculated the approximated distance traveled between consecutive latitude-longitude pairs. It is important to remark that if the maximum time without any location point, which we set to $30$ minutes, was exceeded for some specific hour, the total distance in that interval was considered as a missing value. For the binary case, we preprocessed all the locations in order to estimate the home location, that is, the most usual latitude-longitude pair during nocturnal hours. To decide whether the student was or not at home, we established a maximum distance range of $50$ meters. When one or more location traces show a relative distance to the estimated home location equal or lower than this range, then we assume that at this hour the student was at home. For the hours with no location traces, we considered them also as missing data.

 \begin{figure}[!t]
 	\centering
 	\hspace*{-3.1mm}\includegraphics[width=10.2cm]{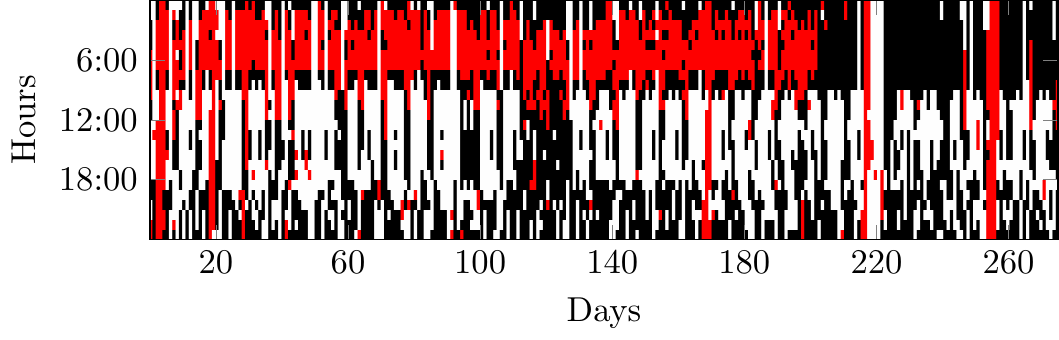}
 	\includegraphics[width=10cm]{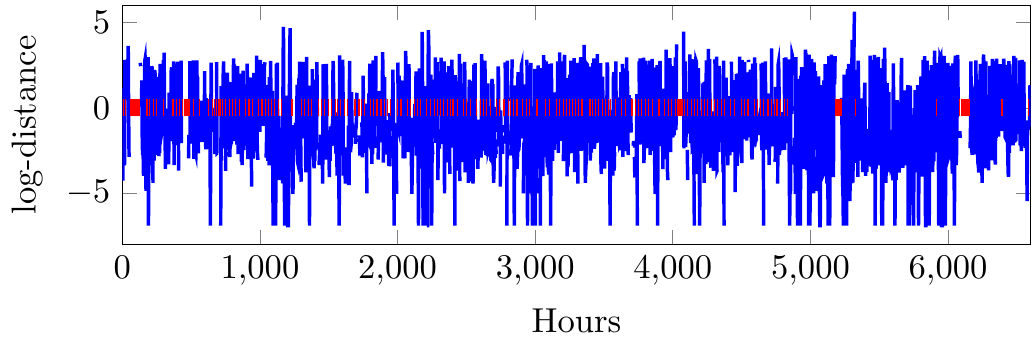}
 	\hspace*{2.5mm}\includegraphics[width=9.7cm]{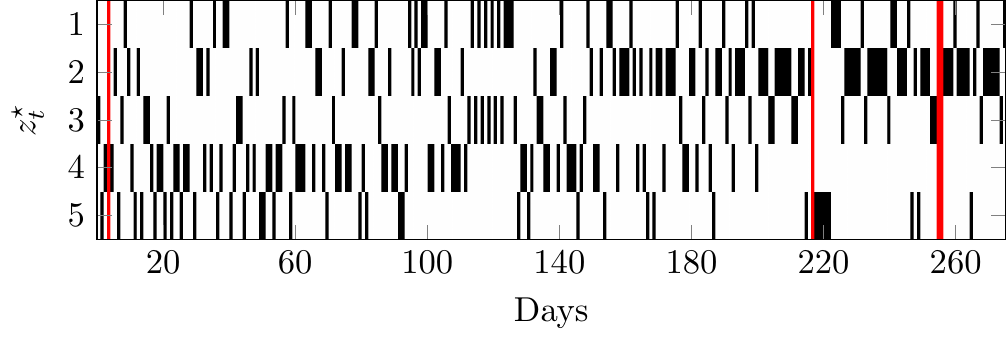}
 	\includegraphics[width=10cm]{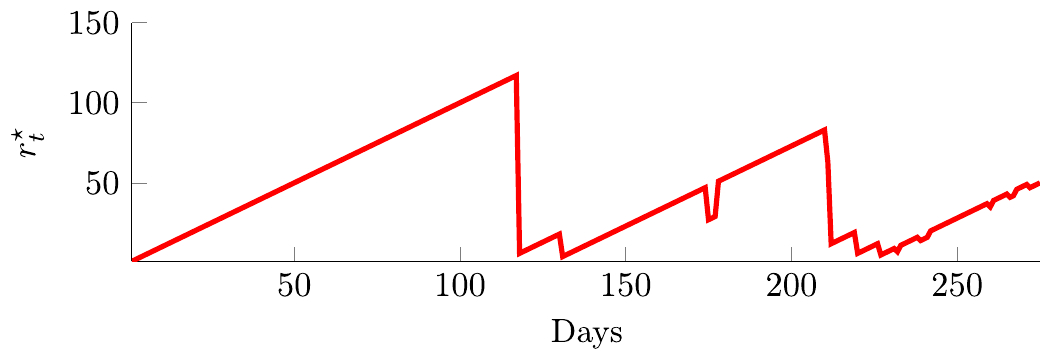}
 	\caption{Results from the experiment with real data. \textbf{First and Second Rows:} Binary and real valued observations per day and hour, respectively. Red traces correspond to missing hours. \textbf{Third Row:} MAP estimates of the latent class indicators, $z^\star_t$, inferred by the PEO model. Red lines are missing latent variables, which correspond to days where all features are lost. \textbf{Fourth Row:} MAP estimates of the run length using the most representative model, i.e., $K=5$ classes.}
 	\label{fig:behavior}
 \end{figure} 

After the preprocessing step, we can see both signals in Figure \ref{fig:behavior} (upper plots), which correspond to the observations $\X_{1:t}$ in the circadian model. For the binary metric, we have $24$-dimensional vectors for each day where at-home (black) and not-at-home (white) indicators are showed, in contrast to missing values (red). Note that it is especially easy to identify patterns and routines in this type of representations \citep{eagle2009eigenbehaviors}. An important detail to remark is that missing traces are mainly concentrated at night. This may occur when smartphones run out of battery during night, which a priori could make sense. Moreover, visualizing segments of diurnal features across the sequence of days, it is interesting since we can infer the routine structure of consecutive working days and weekend periods.

\subsubsection{Detection of Behavioral changes}

The results of the EM algorithm after observing the sequence of heterogeneous observations is shown in Figure \ref{fig:behavior}. Similarly to the previous synthetic experiment, the main objective is to obtain the posterior probabilities $p(\z_{1:t}|\X_{1:t})$. Then, based on the PEO model, we set the point estimates $\z^\star_{1:t}$ as the sequence of observed latent variables, which are shown in the third row of Figure \ref{fig:behavior}. 


Results using $K=5$ classes, this is, a large number of different profiles, show that the second and fourth profiles are the most frequent. Looking at the detected change points, we can see that abrupt transitions are produced by sudden changes in the proportions of latent classes. The hierarchical detector is capturing that in the first partition, the natural parameters $\bm{\pi}_1$ indicate that the fourth and fifth profiles mainly dominate the routine. However, after a small secondary partition given by an unusual combination of classes, the third partition shows a new pattern of behavior where the second profile is the most frequent one. We confirm that the three main CPs detected, respectively at days $t=117, 132$ and $210$ have a ground-truth interpretation for the student under study. Particularly, the short partition between $\text{CP}_1$ and $\text{CP}_2$ (period $\rho=[117, 132]$) corresponds to \textit{pre}-deadline days during holidays, explaining the shift in both mobility and sleep cycles (see Figure \ref{fig:behavior}). Additionally, we identify $\text{CP}_3$ at $t=210$ an its consecutive days, which are actually irregular, as the week of \textit{easter} holidays, when the student was out of the laboratory. One final comment is that, importantly, we find that our method is able to: 1) infer different patterns of routines as, for instance, job/leisure behaviors, 2) CPs appear due to extreme discrepancies between partitions and is robust to outliers (i.e. weekends, uncommon working days) and 3) distinguish (potentially) similar behaviors, as for example, in the case of partitions $\rho_1 = [1, 117]$ and $\rho_3 = [132, 210]$ that corresponds to working epochs but with the difference of being \textit{Autumn} and \textit{Spring} seasons (see patterns of $z^\star_t$ at Figure \ref{fig:behavior}).

\subsubsection{Circadian phenotypes}

\begin{figure}[!t]
	\centering
	\includegraphics[width=7cm,height=13.5cm]{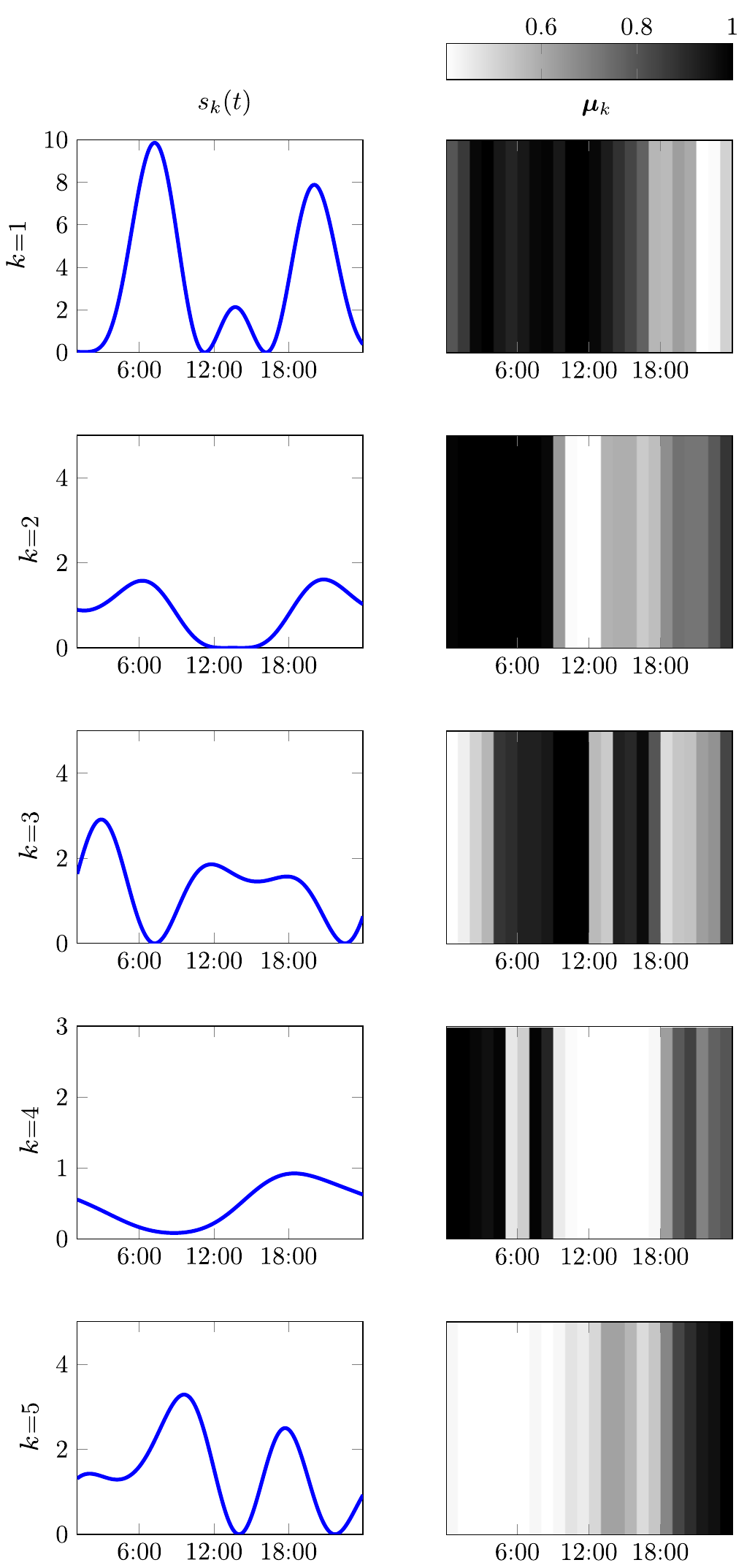}
	\caption{Circadian patterns of mobility variance from both log-distance and binary observations. The results are obtained for $K=5$. \textbf{Left Column:} Periodic functions generated from the squared Fourier series with the estimated hyperparameters. Functions represent the variance in the mobility per hour and type of day. \textbf{Right Column:} Thermal plots of multivariate Bernoulli distribution, i.e., the probability of being at-home during different hours.}
	\label{fig:phenotypes}
\end{figure}

 One of the strengths of the proposed latent model is its ability to identify the $24$-hours periodic patterns from the set of mobility metrics, mainly by estimating the Fourier coefficients of the covariance functions for the traveled distance values. In this process, after fitting all hyperparameters in the maximization step during the inference, we are able to reproduce the patterns that describe how a person usually behaves during each type of day. In the left column of Figure \ref{fig:phenotypes}, we represent the set of (multimodal) periodic functions generated from the squared Fourier series, $s_k(t)$, for every profile, that is, each latent class $z_t = k$. These functions correspond to the variance estimated per hour for each type of behavioral day. Finally, we plot in the right-hand side column of the same figure the estimated vectors of probability $\bm{\mu}_k$ of the Bernoulli likelihood, i.e., the probability of being at home for every type of day at different hourly frames. 
 
The input dependent mapping $s_k(t)$ can be understood by the reader as a periodic function that models the mobility fluctuations during the whole day. That is, large values of $s_k(t)$ correspond to patterns with higher variance of displacements and, on the contrary, values near zero mean less variance in the mobility at the given hour for that particular profile. We also want to remark the interesting connection between resting states (at home) and profiles of traveled distance. Despite the fact that binary observations suffer from extremely high percentages of missing values during nocturnal hours, what sometimes may reduce the correlation between both mobility metrics, it is possible to infer if the patient's pattern of mobility variance corresponds or not to places near his/her home.
  
This circadian phenotypes also provide interesting information for psychiatric use. On one hand, the most likely profiles, such as $k=2$ and $k=4$ (see Figure \ref{fig:behavior}, third row), are easily interpreted as routine days, probably working type ones, as can be seen in Figure \ref{fig:phenotypes}. For instance, Profile $2$ has most of its activity at commuting hours. On the other hand, other profiles, such as $k=3$ or $k=5$, may be considered as leisure patterns since the activity is distributed more evenly during the day. This sort of analysis is particularly important because it demonstrates that heterogeneous circadian models are able to capture the circadian cycles of individuals during their daily life. Actually, this interpretation suggested the nomenclature \emph{type of day}.


\section{Conclusions}

In this paper we have proposed a novel generalization of the Bayesian change-point detection algorithm \citep{adams2007bayesian} to handle heterogeneous high-dimensional observations with unknown periodic structure. The main contribution is to introduce a hierarchical model for the underlying probabilities of the detector. 

We have also presented a new probabilistic approach for detecting change points from sequences with multiple missing values. This solution can be used in any variation of the BOCPD method \citep{saatcci2010gaussian,turner2009adaptive,turner2013online}, and is specially useful when using latent variable models. In addition, we have also proposed a model to capture the periodic structure of the data, mainly focusing on the circadian rhythm in human behavior characterization. 


The code is publicly available in the repository \texttt{https://github.com/pmorenoz/HierCPD} and it is fully written in Python. Thus, we expect the hierarchical detector to be utilised by researchers to detect change points for any possible latent variable model. This package includes the heterogeneous circadian mixture model, as well as the EM inference and the periodic non-stationary covariance functions.




\acks{The authors would like to thank Luca Martino for his useful comments and discussions. This work has been partly supported by the Ministerio de Ciencia, Innovaci{\'o}n y Universidades under grant TEC2017-92552-EXP (aMBITION), by the Ministerio de Ciencia, Innovaci{\'o}n y Universidades, jointly with the European Commission (ERDF), under grants TEC2015-69868-C2-1-R (ADVENTURE) and TEC2017-86921-C2-2-R (CAIMAN), and by The Comunidad de Madrid under grant Y2018/TCS-4705 (PRACTICO-CM). The work of P. Moreno-Mu\~noz has been supported by FPI grant BES-2016-077626.}


\newpage
\appendix

\section*{Appendix A. Hierarchical BOCPD Algorithm}

The recursive decomposition of the joint probability distribution $p(r_t,\mathbf{Z}_{1:t},\mathbf{X}_{1:t}, \bm{\theta}_t)$ is based on the original derivation of \citet{adams2007bayesian} and can obtained as follows. \\

The joint distribution $p(r_t,\mathbf{Z}_{1:t},\mathbf{X}_{1:t},\bm{\theta}_{t})$ can be easily expanded by marginalizing over all values of the previous run length $r_{t-1}$, that is,
\begin{eqnarray}
p(r_t,\mathbf{Z}_{1:t},\mathbf{X}_{1:t},\bm{\theta}_{t}) &=& \sum_{r_{t-1}}p(r_t,r_{t-1},\mathbf{Z}_{1:t},\mathbf{X}_{1:t},\bm{\theta}_t) \nonumber\\
&=& \sum_{r_{t-1}}p(r_t,r_{t-1}, \z_t,\mathbf{Z}_{1:t-1},\mathbf{x}_t,\mathbf{X}_{1:t-1}, \bm{\theta}_t),
\end{eqnarray}
where we have separated $\X_{1:t}$ and $\Z_{1:t}$ to simplify the derivation. The last term can be rewritten as
\begin{equation}
p(r_t,r_{t-1},\z_t,\mathbf{Z}_{1:t-1},\mathbf{x}_t,\mathbf{X}_{1:t-1}, \bm{\theta}_t) = p(r_t,\z_t,\mathbf{x}_t, \bm{\theta}_t|r_{t-1},\mathbf{Z}_{1:t-1},\mathbf{X}_{1:t-1})p(r_{t-1},\mathbf{Z}_{1:t-1},\mathbf{X}_{1:t-1}).
\end{equation}
where the right-hand side term is
\begin{equation}
  \label{eq:previous_marginal_joint}
 p(r_{t-1},\mathbf{Z}_{1:t-1},\mathbf{X}_{1:t-1}) = \int p(r_{t-1},\mathbf{Z}_{1:t-1},\mathbf{X}_{1:t-1}, \bm{\theta}_{t-1}) d\bm{\theta}_{t-1},
 \end{equation} 
 with $p(r_{t-1},\mathbf{Z}_{1:t-1},\mathbf{X}_{1:t-1}, \bm{\theta}_{t-1})$ being the factorized joint probability distribution at $t-1$.

Since the current run length $r_t$ is only conditioned by its previous value $r_{t-1}$, \eqref{eq:previous_marginal_joint} can be written down as
\begin{equation}
p(r_t,\mathbf{z}_{1:t},\mathbf{X}_{1:t}, \bm{\theta}_t) = \sum_{r_{t-1}}p(r_t|r_{t-1})p(z_t,\mathbf{x}_t, \bm{\theta}_t|r_{t-1},\mathbf{Z}_{1:t-1},\mathbf{X}_{1:t-1})p(r_{t-1},\mathbf{z}_{1:t-1},\mathbf{X}_{1:t-1}),
\end{equation}
where $p(r_t|r_{t-1})$ is the change-point prior (see Section \ref{sec:hierarchical}). Note that useless conditioned variables have been omitted. At the same time, we may decompose
\begin{equation}
p(z_t,\mathbf{x}_t,\bm{\theta}_t|r_{t-1},\mathbf{Z}_{1:t-1},\mathbf{X}_{1:t-1}) = p(\mathbf{x}_t|\z_t)p(z_t|\bm{\theta}_t)p(\bm{\theta}_t|r_{t-1},\mathbf{Z}_{1:t-1}),
\end{equation}
where we have taken into account that $\x_t$ is only conditioned by its latent representation $\z_t$, which is modeled by the likelihood term $p(\z_t|\bm{\theta}_t)$ given the posterior distribution $p(\bm{\theta}_t|r_{t-1},\mathbf{Z}_{1:t-1})$ on the parameters.

The resulting recursive expression is
\begin{equation}
p(r_t,\mathbf{z}_{1:t},\mathbf{X}_{1:t}, \bm{\theta}_t) = \sum_{r_{t-1}}p(r_t|r_{t-1})p(\mathbf{x}_t|\z_t)p(\z_t|\bm{\theta}_t)p(\bm{\theta}_t|r_{t-1},\mathbf{Z}_{1:t-1})p(r_{t-1},\mathbf{z}_{1:t-1},\mathbf{X}_{1:t-1}),
\end{equation}
and can be calculated sequentially at each time step $t$.

\section*{Appendix B. Heterogeneous Circadian Mixture Model}

\subsubsection*{Gaussian Likelihood with Missing Data}

The Gaussian distribution requires to handle missing or partial observations in this problem. Thus, based on \citet{ghahramani1994supervised}, the likelihood is as follows
\begin{align}
&p(\mathbf{x}^{\text{real}}_i|\bm{\theta}_k) = p(\mathbf{x}^{o,\text{real}}_i, \mathbf{x}^{m,\text{real}}_i|\bm{\theta}_k) =  \mathcal{N}\Big(\begin{bmatrix}\mathbf{y}^o_i\\\mathbf{y}^m_i\end{bmatrix}\Big|\begin{bmatrix}\bm{0}^{o}\\\bm{0}^{m}\end{bmatrix}, \begin{bmatrix}\bm{\Sigma}_k^{oo} & \bm{\Sigma}_k^{om}\\ \bm{\Sigma}_k^{mo} & \bm{\Sigma}_k^{mm} \end{bmatrix}\Big),
\end{align}
where the blocks of the covariance matrix are given by
\begin{align}
[\bm{\Sigma}^{oo}_k]_{t^{o},t^{o'}} &= g_k(t^{o},t^{o'}) + \sigma_t^2 \cdot \mathbb{I}[t^{o} = t^{o'}], & [\bm{\Sigma}^{mm}_k]_{t^{m},t^{m'}} &= g_k(t^{m},t^{m'}) + \sigma_t^2 \cdot \mathbb{I}[t^{m} = t^{m'}]\\
[\bm{\Sigma}^{mo}_k]_{t^{m},t^{o'}} &= g_k(t^{m},t^{o'}),  &[\bm{\Sigma}^{om}_k]_{t^{o},t^{m'}} &= g_k(t^{o},t^{m'}),
\end{align}
where $t^{o}$ is the time index of an observed variable and $t^{m}$ is the time index of a missing variable. 

%

\subsubsection*{Expected Complete Heterogeneous Log-Likelihood}

The expectation of the complete log-likelihood, which we denote by $\mathcal{Q}$, can be obtained as
\begin{align}
\mathcal{Q}&= \mathbb{E}_{\mathbf{z}_{1:t}, \mathbf{X}^m_{1:t}}[\mathcal{L}_{\z, \bm{\phi}}]\nonumber\\
&=\sum_{i=1}^{t}\sum_{k=1}^{K}\mathbb{E}_{\mathbf{z}_i}\Big[\mathbb{I}\{z_i=k\}\Big]\log \pi_k + \sum_{i=1}^{t}\sum_{k=1}^{K} \mathbb{E}_{\mathbf{z}_i}\Big[\mathbb{I}\{z_i=k\}\Big] \mathbb{E}_{\mathbf{x}^m} \Big[ \log p(\mathbf{x}^o_i, \mathbf{x}^m_i|\bm{\theta}_k)\Big]. 
\end{align}
Substituting $p(\mathbf{x}^o_i, \mathbf{x}^m_i|\bm{\theta}_k)$ by the Bernoulli-Gaussian mixture, we obtain
\begin{align}
\mathcal{Q} = &\sum_{i=1}^{t}\sum_{k=1}^{K}\mathbb{E}_{\mathbf{z}}\Big[\mathbb{I}\{z_i=k\}\Big]\log \pi_k + \sum_{i=1}^{t}\sum_{k=1}^{K}\mathbb{E}_{\mathbf{z}}\Big[ \mathbb{I}\{z_i=k\}\Big] \Big\{-\frac{D}{2}\log(2\pi) - \frac{1}{2}\log\left(|\bm{\Sigma}_k|\right) \nonumber\\
&- \frac{1}{2} \left(\mathbf{x}^{\text{real},o}_i \right)^\top \left(\bm{\Sigma}_k^{oo} \right)^{-1}\mathbf{x}_i^{\text{real},o}  - \frac{1}{2}\mathbb{E}_{\mathbf{x}^{\text{real},m}}\Big[\mathbf{x}^{\text{real},m}_i\Big]^\top \left(\bm{\Sigma}_k^{,mo}\right)^{-1} \mathbf{x}_i^{\text{real},o} \nonumber 
\\ &- \frac{1}{2}\left(\mathbf{x}^{\text{real},o}_i \right)^\top \left(\bm{\Sigma}_k^{om}\right)^{-1}\mathbb{E}_{\mathbf{x}^{\text{real},m}}\Big[\mathbf{x}^{\text{real},m}_i\Big] - \frac{1}{2}\mathbb{E}_{\mathbf{x}^{\text{real},m}}\Big[\mathbf{x}^{\text{real},m}_i\Big]^{\top} \left(\bm{\Sigma}_k^{mm}\right)^{-1} \mathbb{E}_{\mathbf{x}^{\text{real},m}}\Big[\mathbf{x}^{\text{real},m}_i\Big] \nonumber \\
& + \frac{1}{2}\text{tr}(\bm{\Sigma}_k^{-1}\text{Cov}(\mathbf{x}^{\text{real},m}_i)) + \mathbb{E}_{\mathbf{x}^{\text{bin},m}}\Big[\mathbf{x}^{\text{bin},m}_i\Big] \log\bm{\mu}_k^m + \left(1 - \mathbb{E}_{\mathbf{x}^{\text{bin},m}}\Big[\mathbf{x}^{\text{bin},m}_i\Big]\right)\log( 1 - \bm{\mu}_k^m)\nonumber\\
&+ \mathbf{x}^{\text{bin},o}_i \log\bm{\mu}_k^o + \left(1 - \mathbf{x}^{\text{bin},o}_i \right)\log( 1 - \bm{\mu}_k^o)\Big\}. 
\end{align}

\section*{Appendix C. Circadian Model Derivatives}

\subsubsection*{Derivatives of the Heterogeneous Log-Likelihood}


Let us denote the set of hyperparameters of the periodic non-stationary kernel $g_k(t,t')$ as $\bm{\psi}_k = [\ell_k, \mathbf{a}_k^\top, \mathbf{b}_k^\top]^\top$. Additionally, we refer to all variables $\widetilde{\x}$ as the ones whose missing values $\widetilde{\x}^{m}$ have been replaced by the expected ones at each E-step, similarly to the convention adopted in \citet{ghahramani1994supervised}. The derivatives of $\mathcal{Q}$ w.r.t. $\bm{\psi}_k$ and $\bm{\sigma}$ are respectively 
\begin{align}
\frac{\partial \mathcal{Q}}{\partial \bm{\psi}_k} &= \frac{\partial}{\partial \bm{\psi}_k}\left[ - \frac{1}{2}\sum_{i=1}^{t}\sum_{k=1}^{K}r_{ik}^o\log(|\bm{\Sigma}_k|) - \frac{1}{2}\sum_{i=1}^{t}\sum_{k=1}^{K}r_{ik}^o\widetilde{\mathbf{x}}_i^{\text{real}\top}\bm{\Sigma}^{-1}_k\widetilde{\mathbf{x}}^{\text{real}}_i - \frac{1}{2}\sum_{i=1}^{t}\sum_{k=1}^{K}r_{ik}^o\text{tr}(\bm{\Sigma}_k^{-1}\text{Cov}(\mathbf{x}^{m}_i))\right] \nonumber\\
&= \frac{1}{2}\sum_{i=1}^{t}r_{ik}^o \text{tr}\left(\left(\bm{\alpha}\bm{\alpha}^\top - \bm{\Sigma}^{-1}_k\right)\frac{\partial \bm{\Sigma}_k}{\partial \bm{\psi}_k}\right) + \frac{1}{2}\sum_{i=1}^{N}r_{ik}^o \text{tr}\left(\left(\bm{\Sigma}^{-1}_k \mathbf{A}^{\text{old}}_k \bm{\Sigma}^{-1}_k\right)\frac{\partial \bm{\Sigma}_k}{\partial \bm{\psi}_k}\right),
\end{align}
and
\begin{align}
\frac{\partial \mathcal{Q}}{\partial \bm{\sigma}} &= \frac{\partial}{\partial \bm{\sigma}}\left[ - \frac{1}{2}\sum_{i=1}^{t}\sum_{k=1}^{K}r_{ik}^o\log(|\bm{\Sigma}_k|) - \frac{1}{2}\sum_{i=1}^{t}\sum_{k=1}^{K}r_{ik}^o\widetilde{\mathbf{x}}_i^{\text{real}\top}\bm{\Sigma}^{-1}_k\widetilde{\mathbf{x}}^{\text{real}}_i - \frac{1}{2}\sum_{i=1}^{t}\sum_{k=1}^{K}r_{ik}^o\text{tr}\left(\bm{\Sigma}_k^{-1}\text{Cov}(\mathbf{x}^{m}_i)\right)\right]\nonumber\\
&= \frac{1}{2}\sum_{i=1}^{t} \sum_{k=1}^{K} r_{ik}^o \text{tr}\left(\left(\bm{\alpha}\bm{\alpha}^\top - \bm{\Sigma}^{-1}_k\right)\frac{\partial \bm{\Sigma}_k}{\partial \bm{\sigma}}\right) + \frac{1}{2}\sum_{i=1}^{t} \sum_{k=1}^{K} r_{ik}^o \text{tr}\left(\left(\bm{\Sigma}^{-1}_k \mathbf{A}^{\text{old}}_k \bm{\Sigma}^{-1}_k\right)\frac{\partial \bm{\Sigma}_k}{\partial \bm{\sigma}}\right),
\end{align}
where $\bm{\alpha} = \bm{\Sigma}^{-1}_k \widetilde{\mathbf{x}}_i,$ with
\begin{align}
\widetilde{\mathbf{x}}^{{\text{real}}}_i &= \begin{bmatrix} \mathbf{x}^{{\text{real}},o}_i\\\mathbb{E}_{\mathbf{x}^{{\text{real}},m}}[\mathbf{x}^{{\text{real}},m}_i]\end{bmatrix}, &
\mathbf{A}^{\text{old}}_k &= \begin{bmatrix}\bm{0}^{oo} & \bm{0}^{om}\\ \bm{0}^{mo} & \text{Cov}^{\text{old}}(\mathbf{x}^{m}_k) \end{bmatrix}.
\end{align}
Here, we have used that 
\begin{equation}
\frac{\partial \bm{\Sigma}_k}{\partial \bm{\psi}_k} = \frac{\partial \mathbf{K}_k}{\partial \bm{\psi}_k},
\end{equation}
and
\begin{equation}
\frac{\partial \bm{\Sigma}_k}{\partial \bm{\sigma}} =  \frac{\partial \mathbf{D}}{\partial \bm{\sigma}}.
\end{equation}

\subsubsection*{Derivatives of the Periodic Non-stationary Kernel}
\label{app:derivatives}


In this appendix, we present the derivatives w.\,r.\,t.\ $\ell_k$, $\mathbf{a}_k$ and $\mathbf{b}_k$, which are given by
\begin{align}
\frac{\partial g_{k}(t,t')}{\partial \ell_k} &= s_k(t)^2 s_k(t')^2\exp \left(-\frac{2\sin^2(\pi(t-t')/T)}{\ell_k^2}\right)\left(\frac{4\sin^2(\pi(t-t')/T)}{\ell_k^3}\right)\nonumber\\
\frac{\partial g_{k}(t,t')}{\partial a_{0k}} &= s_k(t) s_k(t')\exp \left(-\frac{2\sin^2(\pi(t-t')/T)}{\ell_k^2}\right)\left(s_k(t)+s_k(t')\right)\nonumber\\
\frac{\partial g_{k}(t,t')}{\partial a_{nk}} &= 2\exp \left(-\frac{2\sin^2(\pi(t-t')/T)}{\ell_k^2}\right)s_k(t)s_k(t')\left(\cos \left(\frac{2\pi n}{T}t\right)s_k(t') + \cos \left(\frac{2\pi n}{T}t'\right)s_k(t)\right)\nonumber\\
\frac{\partial g_{k}(t,t')}{\partial b_{nk}} &= 2\exp \left(-\frac{2\sin^2(\pi(t-t')/T)}{\ell_k^2}\right)s_k(t)s_k(t')\left(\sin \left(\frac{2\pi n}{T}t\right)s_k(t') + \sin \left(\frac{2\pi n}{T}t'\right)s_k(t)\right)\nonumber
\end{align}

\section*{Appendix D. On Experiments and EM Algorithm Details}
\label{appendix:initialization}

 In the repository \texttt{https://github.com/pmorenoz/HierCPD/}, the \textsc{Python} version of our code is publicly available, for reproducibility of both the hierarchical CPD and circadian heterogeneous models. In this case, we make use of the \textit{Paramz} optimization module, nested into the \textsc{GPy} software library, specially indicated for probabilistic learning simulations.\\
 
  \noindent\textbf{EM algorithm simulations:} We initialize the parameters $\bm{\pi}_k$ using a \text{Dirichlet} distribution with the form $\text{Dir}(c\bm{1})$, where $c=50$ and $\bm{1}$ is a vector of ones, with size equal to $K$. All parameters $\bm{\mu}_k$ for Bernoulli distributions are initially sampled from an $\text{Uniform}(0,1)$ and similarly, every component in $\bm{\sigma}$. The hyperparameters $\mathbf{a}_k, \mathbf{b}_k, \sigma_{ak}$ and $\ell_k$ of non-stationary periodic kernels $g_k(t,t')$ are initialized using an $\text{Uniform}(2,1)$ in the case of $\sigma_{ak}$ and $\ell_k$, and for the  Fourier series coefficients, $\mathbf{a}_k$ and $\mathbf{b}_k$, we sampled from $\mathcal{N}(0,1)$.
  
  For experiments of our circadian heterogeneous mode, we ran multiple simulations using $5$ different initializations, a maximum number of $10$ optimization evaluations per M-step and a stopping/convergence criteria dependent on $\epsilon=250$ for the function $\mathcal{Q}$. The number of Fourier coefficients was set to $C=3$ and we always considered a circadian period $D=24$ in $\eqref{eq:fourier}$.
  
   \noindent\textbf{Time efficiency:} Our hierarchical CPD method behaves differently, in terms of computing time, for the CP detector and the heterogeneous mixture model. While the first one updates analytically the posterior distribution over the run length, the second needs to optimize parameters w.\,r.\,t.\ the complete log-likelihood function $\mathcal{Q}$. Particularly, in the CPD case, the computational cost is the same one as in \citet{adams2007bayesian}, which could be problematic if the number of time steps is larger than $t=10^3$. This could force us to use pruning strategies as in \citet{turner2009adaptive}, but was not necessary in our application. In the second case, the mixture model includes a sequential optimization step whose computation time is given in terms of tens of minutes (10-40) using a standard computer. Note that this also depends of the number of initializations, the maximum number of evaluations allowed and the convergence criteria.

\vskip 0.2in
\bibliography{paper_hiercpd}

\end{document}